\documentclass{article}
\usepackage{times}

\usepackage[accepted]{icml2025}

\usepackage[english]{babel}
\usepackage{comment}
\usepackage{natbib}
\usepackage{amsmath}
\usepackage{amsthm}
\usepackage{amsfonts}
\usepackage{amssymb}
\usepackage{stmaryrd}
\usepackage{bm}
\usepackage{commath}
\usepackage{dsfont}
\usepackage{pifont}
\usepackage{microtype}
\usepackage{graphicx}
\usepackage{subfigure}
\usepackage{wrapfig}
\usepackage{multirow}
\usepackage{multido}
\usepackage{makecell}
\usepackage{booktabs}
\usepackage{nicefrac}
\usepackage{tikz}
\usepackage{pgfplots}
\usepackage{xcolor}         
\usepackage[separate-uncertainty=true]{siunitx}
\usepackage[colorlinks=true, allcolors=blue]{hyperref}
\usepackage[capitalize, noabbrev]{cleveref}

\usepackage{amsmath,amsfonts,bm}

\def\eqref#1{equation~\ref{#1}}

\def\1{\bm{1}}

\def\rx{{\textnormal{x}}}

\def\rvv{{\mathbf{v}}}
\def\rvw{{\mathbf{w}}}
\def\rvx{{\mathbf{x}}}
\def\rvy{{\mathbf{y}}}
\def\rvz{{\mathbf{z}}}

\def\vf{{\bm{f}}}
\def\vg{{\bm{g}}}

\DeclareMathAlphabet{\mathsfit}{\encodingdefault}{\sfdefault}{m}{sl}
\SetMathAlphabet{\mathsfit}{bold}{\encodingdefault}{\sfdefault}{bx}{n}

\newcommand{\E}{\mathbb{E}}
\newcommand{\Ls}{\mathcal{L}}

\newtheorem{theorem}{Theorem}
\newtheorem{lemma}{Lemma}

\newtheorem{corollary}{Corollary}

\newtheorem{observation}{Finding}
\theoremstyle{definition}

\Crefname{assumption}{Assumption}{Assumptions}
\Crefname{observation}{Finding}{Findings}
\makeatletter
    \newtheorem*{rep@theorem}{\rep@title}
    \newcommand{\newreptheorem}[2]{%
    \newenvironment{rep#1}[1]{%
    \def\rep@title{#2 \ref{##1}}%
    \begin{rep@theorem}}%
    {\end{rep@theorem}}}
\makeatother
\newreptheorem{theorem}{Theorem}
\newreptheorem{lemma}{Lemma}
\newreptheorem{corollary}{Corollary}

\bibpunct{(}{)}{;}{a}{,}{,}
\pgfplotsset{compat=1.16}

\icmltitlerunning{Improving Consistency Models with Generator-Augmented Flows}

\begin{document}

\twocolumn[
\icmltitle{Improving Consistency Models with Generator-Augmented Flows}



\icmlsetsymbol{equal}{*}

\begin{icmlauthorlist}
\icmlauthor{Thibaut Issenhuth}{criteo}
\icmlauthor{Sangchul Lee}{kist}
\icmlauthor{Ludovic Dos Santos}{criteo}
\icmlauthor{Jean-Yves Franceschi}{criteo}
\icmlauthor{Chansoo Kim}{kist,ust}
\icmlauthor{Alain Rakotomamonjy}{criteo,litis}
\end{icmlauthorlist}


\icmlaffiliation{criteo}{Criteo AI Lab, Paris, France}
\icmlaffiliation{kist}{AI, Information and Reasoning (AI/R) Laboratory, Korea Institute of Science and Technology}
\icmlaffiliation{ust}{AI and Robot Department, University of Science and Technology, Korea}
\icmlaffiliation{litis}{LITIS, Univ Rouen-Normandie}
\icmlcorrespondingauthor{Thibaut Issenhuth, Chansoo Kim}{t.issenhuth@criteo.com; eau@ust.ac.kr}

\icmlkeywords{Machine Learning, ICML}

\vskip 0.3in
]

\printAffiliationsAndNotice 

\begin{abstract}
Consistency models imitate the multi-step sampling of score-based diffusion in a single forward pass of a neural network.
They can be learned in two ways: consistency distillation and consistency training. The former relies on the true velocity field of the corresponding differential equation, approximated by a pre-trained neural network.
In contrast, the latter uses a single-sample Monte Carlo estimate of this velocity field.
The related estimation error induces a discrepancy between consistency distillation and training that, we show, still holds in the continuous-time limit.
To alleviate this issue, we propose a novel flow that transports noisy data towards their corresponding outputs derived from a consistency model.
We prove that this flow reduces the previously identified discrepancy and the noise-data transport cost.
Consequently, our method not only accelerates consistency training convergence but also enhances its overall performance. The code is available at: \href{https://github.com/thibautissenhuth/consistency_GC}{ 
github.com/thibautissenhuth/consistency\_GC}.


\end{abstract}

\section{Introduction}

\begin{figure*}
    \centering
    {
    \subfigure[PF-ODE (IC).]{\includegraphics[width=0.46\textwidth]{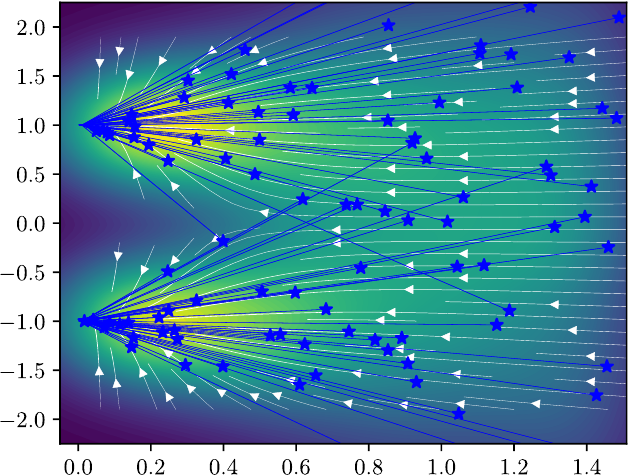}}
    \hspace{0.8cm}
    \subfigure[Generator-Augmented Flows (GC).]{\includegraphics[width=0.46\textwidth]{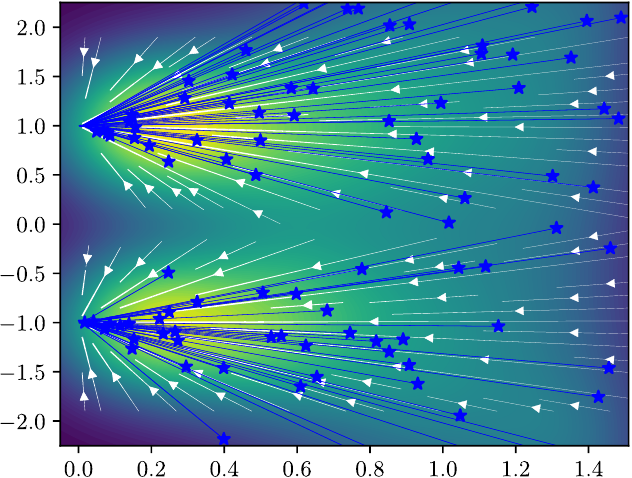}}
    }    
    \caption{
        Comparison of the probability flow ODE (PF-ODE) and generator-augmented flows (GC): target data is a mixture of two Dirac delta functions, and GC is computed with a closed-form generator. In the background, we observe the density of probability paths. White arrows are ODE trajectories associated to the velocity field. Blue lines are sample paths from IC in (a) and from GC in (b). Trajectories start from random intermediate points \textcolor{blue}{\ding{72}}. On this example, GC sample paths appear more aligned to the velocity field.  
        \label{fig:gaussian_viz}
    }
\end{figure*}

A large family of diffusion \citep{ho2020denoising}, score-based \citep{song2021scorebasedSDEs,karras2022edm}, and flow models \citep{liu2023flow,lipman2023flow} have emerged as state-of-the-art generative models for image generation. Since they are costly to use at inference time --~requiring several neural function evaluations~--, many distillation techniques have been explored \citep{salimans2022progressive, meng2023distillation, sauer2023adversarial}. One of the most remarkable approach is \emph{consistency models} \citep{song23consistency, song2023improved}. Consistency models lead to high-quality one-step generators, that can be trained either by distillation of a pre-trained velocity field (\emph{consistency distillation}), or as standalone generative models (\emph{consistency training)} by approximating the velocity field through a one-sample Monte Carlo estimate. 

The corresponding estimation error naturally induces a discrepancy between consistency distillation and training.
While \citet{song23consistency} hinted that it would resolve in the continuous-time limit, we show that this discrepancy persists in both the gradients and values of the loss functions.
Interestingly, this discrepancy vanishes when the difference between the target velocity field and its Monte-Carlo approximation approaches zero.
However, this is not the case with the independent coupling (IC) between data and noise used to construct the standard estimate.
It is unclear how to improve this one-sample estimate without access to the true underlying diffusion model.


The approach we adopt in this paper to alleviate this issue involves altering the velocity field --~thereby changing the target flow~-- to reduce the variance of its one-sample estimator. One possible solution to this problem is to resort to optimal transport (OT) to learn on a deterministic coupling. OT has been succesfully adopted in diffusion \citep{li2024immiscible}, consistency \citep{dou2024a}, and flow matching \citep{pooladian2023multisample} models.  
However, due to the prohibitive cubic complexity of OT solvers (\textit{e.g.}\ Hungarian matching algorithm), such methods need to be applied at the minibatch level.
This incurs an OT approximation error \citep{fatras2021minibatch, sommerfeld2019optimal} and stochasticity of the data-noise coupling, thus not solving the consistency training issue.

{In our approach, we propose to use the consistency model, assumed to be an approximation of the target diffusion flow, to construct {additional} trajectories. The consistency model serves as a proxy to reduce the expected deviation between the velocity field and its estimator. }
More precisely, from an intermediate point computed from an IC, we let the consistency model predict the corresponding endpoint, supposedly close to the data distribution.
This predicted endpoint is coupled to the same original noise vector, defining a generator-augmented coupling (GC).
We show empirically that the resulting generator-augmented flow presents compelling properties for training consistency models, in particular a reduced deviation between the velocity field and its estimator, and decreased transport costs ~-- as supported by theoretical and empirical evidence.
This can be observed in \cref{fig:gaussian_viz}.
From this, we derive practical algorithms to train consistency models with generator-augmented flows, leading to improved performance and faster convergence compared to standard and OT-based consistency models.

Let us summarize our contributions below.
\begin{itemize}
    \item 
    We prove that in the continuous-time limit consistency training and consistency distillation loss function converge to different values and we provide a closed-form expression of this discrepancy.
    
    \item We propose a novel type of flows that we denote \textit{generator-augmented flows}. It relies on generator-augmented coupling (GC) that can be used to train a consistency model. 
    \item We provide theoretical and empirical insights into the advantages of GC. We show that generator-augmented flows {have smaller discrepancy  
    to consistency distillation  than IC consistency training,}  
    and that they reduce data-noise transport costs.
    \item We derive practical ways to train consistency models with GC. Our approach based on a joint learning strategy leads to faster convergence and improves the performance compared to the base model and OT-based approaches on image generation benchmarks.
\end{itemize}

\paragraph{Notation.} 
We consider an empirical data distribution $p_\star$ and a noise distribution $p_z$ (\textit{e.g.}\ Gaussian), both defined on $\mathds{R}^d$.
We denote by $q$ a joint distribution of samples from $p_\star$ and $p_z$. We equip $\mathds{R}^d$ with the dot product $\langle \rvx, \rvy \rangle = \rvx^\top \rvy$ and write $\|\rvx\| = \langle \rvx, \rvx \rangle^{1/2}$ for the Euclidean norm of $\rvx$.
We use a distance function $\mathcal{D} \colon \mathds{R}^d \times \mathds{R}^d \to [0, \infty)$ to measure the distance between two points from $\mathds{R}^d$.
$\mathrm{sg}$ denotes the stop-gradient operator.

In consistency models, we consider diffusion processes of the form $\rvx_t = \rvx_\star + \sigma_t \rvz$, where $\rvx_\star \sim p_\star$, $\rvz \sim p_z$, and $\sigma_t$ is monotonically increasing for $t \in [0, T]$, where $T\in \mathds{R}_+$. 
We denote the distribution of $\rvx_t$ by $p(\rvx_t)$, or simply $p_t$. Conditional distributions or finite-dimensional joint distributions of $\rvx_t$'s are denoted similarly.
When considering a discrete formulation with $N$ intermediate timesteps, we denote the intermediate points as $\rvx_{t_i} = \rvx_\star + \sigma_{t_i} \rvz$, where $t_i$ is strictly increasing for $i \in \{0,\ldots,N\}$, with $t_0 = 0$ and $t_N = T$. The values of $\sigma_{0}$ and $\sigma_{T}$ are chosen to be sufficiently small and large, respectively, so that $p_{0} \approx p_\star$ and $p_{T} \approx \mathcal{N}(0, \sigma_T^2 \mathbf{I})$. 

\section{Consistency Distillation Versus Training}

In this section, we provide the required background on diffusion and consistency models (\cref{sec:prelim_diff,sec:prelim_consistency}), then discuss the discrepancy between consistency distillation and consistency training (\cref{sec:prelim_discrepancy}) which we theoretically characterize in continuous-time.

\subsection{Flow and Score-Based Diffusion Models}
\label{sec:prelim_diff}

Score-based diffusion models \citep{ho2020denoising, song2021scorebasedSDEs} can generate data from noise via a multi-step process consisting in numerically solving either a stochastic differential equation (SDE), or equivalently an ordinary differential equation (ODE). 
Although SDE solvers generally exhibit superior sampling quality, ODEs have desirable properties. Most notably, they define a deterministic mapping from noise to data. Recently, \citet{liu2023flow} and \citet{lipman2023flow} generalize diffusion to flow models, which are defined by the following probability flow ODE (PF-ODE):
\begin{equation}
    \dif \rvx = \rvv_t(\rvx) \dif t,
\end{equation}
where $\rvv_t(\rvx) = \E[\dot{\rvx}_t | \rvx_t=\rvx]$ is the velocity field. Note that $\dot{\rvx}_t$ is defined as the random variable $\dot{\rvx}_t = \frac{\dif(\rvx_\star + \sigma_t \rvz)}{\dif t} = \dot{\sigma}_t \rvz$, and is not to be confused with the time-derivative of the ODE, $\rvv_t$.

In the context of consistency models \citep{song23consistency, song2023improved}, the most common choice is $\rvv_t(\rvx) = - \dot{\sigma}_t \sigma_t \nabla_\rvx \log p_t(\rvx) \dif t $, in particular the EDM formulation \citep{karras2022edm} where $\sigma_t = t$ and thus $\rvv_t(\rvx) = - t \nabla_\rvx \log p_t(\rvx)$. Here, $\nabla_\rvx \log p_t$, \textit{a.k.a.}\ the score function, can be approximated with a neural network $\mathbf{s}_\phi(\rvx, t)$ \citep{vincent2011connection, Song2019}.

\subsection{Consistency Models}
\label{sec:prelim_consistency}

Numerically solving an ODE is costly because it requires multiple expensive evaluations of the velocity function.
To alleviate this issue, \citet{song23consistency} propose training a \emph{consistency model} $\vf_\theta$, which learns the output map of the PF-ODE, \emph{i.e.}\ its flow, such that:
\begin{equation}\label{eq:consistency}
    \vf_\theta(\rvx_t,\sigma_t) = \rvx_0, 
\end{equation}
for all $(\rvx_t,\sigma_t) \in \mathds{R}^d \times [\sigma_0,\sigma_T]$ that belong to the trajectory of the PF-ODE ending at $(\rvx_0, \sigma_0)$.

\cref{eq:consistency} is equivalent to \emph{(i)} enforcing the boundary condition $\vf_\theta(\rvx_0, \sigma_0) = \rvx_0$, and \emph{(ii)} ensuring that $\vf_\theta$ has the same output for any two samples of a single PF-ODE trajectory --~the consistency property.
\emph{(i)} is naturally satisfied by the following model parametrization:
\begin{equation}\label{eq:parametrization}
    \vf_\theta(\rvx_{t_i},\sigma_{t_i}) = c_{\text{skip}}(\sigma_{t_i}) \rvx_{t_i} + c_{\text{out}}(\sigma_{t_i}) \bm{F}_\theta(\rvx_{t_i},\sigma_{t_i}),
\end{equation}
where $c_{\text{skip}}(\sigma)=\frac{\sigma_d^2}{\sigma_d^2 + (\sigma - \sigma_0)^2}$, $c_{\text{out}}(\sigma) = \frac{\sigma_d \cdot (\sigma - \sigma_0)}{\sqrt{\sigma_d^2 + \sigma^2}}$, $\sigma^2_d$ the variance of data, and $\bm{F}_\theta$ is a neural network. This ensures $ c_{\text{skip}}(0) = 1$, $ c_{\text{out}}(0) = 0$.
\emph{(ii)} is achieved by minimizing the distance between the outputs of two same-trajectory consecutive samples using the consistency loss:
\begin{multline}
\label{eq:consistency_distill_loss}
    \Ls_{\text{CD}}(\theta)
    = \E_{ q_\mathrm{I}(\rvx_\star, \rvz), p(\rvx_{t_{i+1}} | \rvx_\star, \rvz) } \\
    \quad \Big[ \lambda(\sigma_{t_i}) \mathcal{D}\Big(\mathrm{sg}\bigl(\vf_\theta(\rvx_{t_{i}}^{\Phi}, \sigma_{t_i})\big), \vf_\theta(\rvx_{t_{i+1}},\sigma_{t_{i+1}})\Big)\Big],
\end{multline}
where $(\rvx_\star, \rvz)$ is sampled from the \emph{independent} coupling $q_\mathrm{I}(\rvx_\star, \rvz) = p_\star(\rvx_\star) p_z(\rvz)$, $i$ is an index sampled uniformly at random from $\{0, 1, \ldots, N-1\}$, $\rvx_{t_{i+1}} = \rvx_\star + \sigma_{t_{i+1}}\rvz$, and $\rvx_{t_i}^{\Phi}$ is computed by discretizing the PF-ODE with the Euler scheme as follows:
\begin{equation} \label{eq:score_update}
    \rvx^{\Phi}_{t_{i}} = \Phi( \rvx_{t_{i+1}},t_{i+1}) = \rvx_{t_{i+1}} + (t_i - t_{i+1})  \rvv_{t_{i+1}} (\rvx_{t_{i+1}}).
\end{equation}
This loss can be used to distill a score model into $\vf_\theta$.

In the case of consistency training, \citet{song23consistency} circumvent the lack of a score function by noting that $\rvv_{t_{i+1}} (\rvx) = \E[\dot{\rvx}_{t_{i+1}} | \rvx_{t_{i+1}} = \rvx]$.
In light of this, its single-sample Monte Carlo estimate $\dot{\rvx}_{t_{i+1}}$ is used instead in \cref{eq:score_update} to replace the intractable $\rvx_{t_i}^{\Phi}$ by $\rvx_{t_i} = \rvx_\star + \sigma_{t_{i}}\rvz$ in the consistency loss:
\begin{multline}
\label{eq:consistency_training_loss}
    \Ls_{\text{CT}}(\theta)
    = \E_{q_\mathrm{I}(\rvx_\star, \rvz), p(\rvx_{t_i}, \rvx_{t_{i+1}} | \rvx_\star, \rvz)} \\
    \quad \Big[ \lambda(\sigma_{t_i}) \mathcal{D}\Big(\mathrm{sg}\bigl(\vf_\theta(\rvx_{t_{i}}, \sigma_{t_i})\big), \vf_\theta(\rvx_{t_{i+1}},\sigma_{t_{i+1}})\Big)\Big].
\end{multline}

\subsection{Discrepancy Between Consistency Training and Distillation and Velocity Field Estimation}
\label{sec:prelim_discrepancy}

Naturally, replacing $\rvv_{t}$ by its single-sample estimate $\dot{\rvx}_t$ makes consistency training deviate from consistency distillation in discrete time.
Still, \citet[Theorems~2 and~6]{song23consistency} suggest that this discrepancy disappears in continuous-time since $\Ls_{\text{CT}}(\theta) = \Ls_{\text{CD}}(\theta) + o(\nicefrac{1}{N})$ and the corresponding gradients are equal in some cases.
{This equality is then used in work of \citet{lu2024simplifying}, concurrent to ours, to train continuous-time consistency models at the cost of an elaborate architectural design.}
Without disproving these results, we find that scaling issues and lack of generality 
soften the claim of a closed gap between consistency training and distillation.

Indeed, we provide in the following theorem a thorough theoretical comparison of $\Ls_{\text{CT}}$ and $\Ls_{\text{CD}}$.
We first prove that they converge to different values in the continuous-time limit.
The difference is captured by a regularization term that depends on the discrepancy between the velocity field and its estimate.
Moreover, we show that the limits of the scaled gradients do not coincide in the general case, except when the (asymptotic) quadratic loss is used. 
The proof, and further discussion on why this discrepancy did not appear in \citet{song23consistency}, can be found in \cref{app:proof_discrepancy}.



\begin{theorem}[\textbf{Discrepancy between consistency distillation and consistency training objectives}]  \label{theorem_R_theta}
    Assume that the distance function is given by $\mathcal{D}(\rvx, \mathbf{y}) = \varphi(\| \rvx - \mathbf{y}\|)$ for a continuous convex function $\varphi : [0, \infty) \to [0, \infty)$ with $\varphi(x) \sim Cx^{\alpha}$ as $x \to 0^+$ for some $C > 0$ and $\alpha \geq 1$, and that the timesteps are equally spaced, i.e., $t_i = \frac{i T}{N}$. Furthermore, assume that the Jacobian $\frac{\partial \vf_\theta}{\partial \rvx}$ does not vanish identically. Then the following assertions hold:
    
        \paragraph{\textit{(i)}} The scaled consistency losses $N^\alpha \Ls_\mathrm{CD}(\theta)$ and $N^\alpha \Ls_\mathrm{CT}(\theta)$ converge as $N \to \infty$. Moreover, the minimization objectives corresponding to these limiting scaled consistency losses are not equivalent, and their difference is given by:
        \begin{equation}
            \lim_{N\to\infty} N^\alpha \left[ \Ls_\mathrm{CT}(\theta) - \Ls_\mathrm{CD}(\theta) \right]
            = CT^{\alpha-1} \mathcal{R}(\theta),
        \end{equation}
        where $\mathcal{R}(\theta)$ is defined by
        \begin{equation}
            \mathcal{R}(\theta)
            = \int_{0}^{T} \lambda(\sigma_t) \E \left[ \left\| \partial_\mathrm{CT}\vf_\theta \right\|^\alpha - \left\| \partial_\mathrm{CD}\vf_\theta \right\|^\alpha \right] \, \dif t
        \end{equation}
        and satisfies $\mathcal{R}(\theta) > 0$, with
        \begin{equation}
            \partial_\mathrm{CT}\vf_\theta
            = \frac{\partial \vf_\theta}{\partial \sigma}(\rvx_t, \sigma_t) \dot{\sigma}_t + \frac{\partial \vf_\theta}{\partial \rvx}(\rvx_t, \sigma_t) \cdot \dot{\rvx}_t,
        \end{equation}
        \begin{equation}
            \partial_\mathrm{CD}\vf_\theta
            = \frac{\partial \vf_\theta}{\partial \sigma}(\rvx_t, \sigma_t) \dot{\sigma}_t + \frac{\partial \vf_\theta}{\partial \rvx}(\rvx_t, \sigma_t) \cdot \rvv_t(\rvx_t).
        \end{equation}
        In particular, if $\alpha=2$,
        \begin{equation}
        \label{eq:regularization_term}
            \mathcal{R}(\theta)
            = \int_{0}^{T} \lambda(\sigma_t) \E \bigg[ \left\| \frac{\partial \vf_\theta}{\partial \rvx}(\rvx_t, \sigma_t) \left( \dot{\rvx}_t - \rvv_t(\rvx_t) \right) \right\|^2 \bigg] \dif t.
        \end{equation}

        \paragraph{\textit{(ii)}} The scaled gradient $N^{\alpha-1} \nabla_\theta \Ls_\mathrm{CD}(\theta)$ and $N^{\alpha-1} \nabla_\theta \Ls_\mathrm{CT}(\theta)$ converge as $N \to \infty$. Moreover, if $\alpha \neq 2$, then their respective limits are not identical as functions of $\theta$:
        \begin{equation}
            \lim_{N\to\infty} N^{\alpha-1} \nabla_\theta \Ls_\mathrm{CT}(\theta)
            \neq \lim_{N\to\infty} N^{\alpha-1} \nabla_\theta \Ls_\mathrm{CD}(\theta).
        \end{equation}
\end{theorem}
This theorem reveals that the optimization problems of consistency training and distillation differ not only in discrete time but also in continuous-time.
It even highlights a discrepancy between, firstly, the limiting gradients in continuous-time --~although they are equal for $\alpha=2$~-- and, secondly, the gradients of the limiting losses, which differ because of $\mathcal{R}(\theta)$, even when $\alpha=2$.

This  analysis shows the importance of employing probability paths whose sample path derivatives $\dot{\rvx}_t$ are aligned with the velocity field $\rvv_t(\rvx_t)$.
In particular, if a diffusion process $\rvx_t$ satisfies $\dot{\rvx}_t = \rvv_t(\rvx_t)$, we have $ \mathcal{R}(\theta)=0$ and equal gradients for all $\alpha \geq 1$.
Hence, for such $\rvx_t$, consistency training and consistency distillation would be reconciled both in discrete time and in the continuous-time limit.

However, it is unclear how to directly improve the single-sample estimation $\dot{\rvx}_t$ of $\rvv_t(\rvx_t)$.
In particular, increasing the number of samples per point $\rvx_t$ to reduce its variance is not tractable, as it requires sampling from the inverse diffusion process $p(\rvx_\star | \rvx_t)$.
Therefore, we adopt an alternative approach to alleviate the discrepancy identified in this section, which involves altering the velocity field --~thereby changing the target flow~-- to reduce the variance of its one-sample estimator.
This approach is reminiscent of recent work tackling the data-noise coupling that we discuss in the following section.



\section{Reducing the Discrepancy with Data-Noise Coupling}
\label{sec:coupling}

\paragraph{Beyond independent coupling (IC).}

From \cref{sec:prelim_consistency}, it appears that $\dot{\rvx}_t$ is computed through an IC $q_\mathrm{I} = p_\star(\rvx
_\star) p_z(\rvz)$ of data and noise, in a similar fashion to flow matching \citep{lipman2023flow, kingma2024understanding}.
Making correlated choices of data and noise beyond IC could then help align $\dot{\rvx}_t$ and $\rvv_t(\rvx_t)$, thereby resolving the discrepancy from the previous section.


The reliance on IC in consistency and flow models is increasingly recognized as a limiting factor. 
Recent advancements suggest that improved coupling mechanisms could enhance both training efficiency and the quality of generated samples in flow matching \citep{liu2023flow, pooladian2023multisample} and diffusion models \citep{li2024immiscible}.
By reducing the variance in gradient estimation, enhanced coupling can accelerate training. 
Additionally, improved coupling could decrease transport costs and straighten trajectories, yielding better-quality samples. In a different context, ReFlow \citep{liu2023flow} leverages couplings provided by the ODE solver in a flow framework, and demonstrates that it reduced transport costs. Moreover, \citet{lee23minimizing} propose to learn an encoder from data to noise, and use this encoder as a way to construct a coupling when training a flow model. 

\paragraph{Couplings based on optimal transport (OT) solvers. }
OT is a particularly appealing solution for our alignment problem.
Indeed, if we consider a quadratic cost and distributions with bounded supports, OT is a no-collision transport map \citep{nurbekyan2020no}, \textit{i.e.} $\rvx_t$ can be sampled by a unique pair of points $(\rvx_\star,\rvz)$.
Thus $\dot{\rvx}_t = \rvv_t(\rvx_t)$, implying $\mathcal{R}(\theta)=0$ in \cref{theorem_R_theta}.
Several approaches have precisely targeted the reduction of transport cost in flow and consistency models. 

\citet{pooladian2023multisample} have more directly explored OT coupling within the framework of flow matching models. They show that deterministic and non-crossing paths enabled by OT with infinite batch size lowers the variance of gradient estimators. Experimentally, they assess the efficacy of OT solvers, such as Hungarian matching and Sinkhorn algorithms, in coupling batches of noise and data points. \citet{dou2024a} have successfully adopted this approach in consistency models, while \citet{li2024immiscible} applied OT to diffusion models.
However, due to the prohibitive cubic complexity of OT solvers, OT has to be applied by minibatch for matching samples $(\rvx_\star,\rvz)$.
Besides an OT approximation error, this incurs the loss of the no-collision property, making $\mathcal{R}(\theta)$ non-zero in real use-cases.
Another line of works using OT tools with score-based models relies on the Schr\"{o}dinger Bridge formulation \citep{de2021diffusion, shi2024diffusion, korotin2024light, tong2024simulation}, which has mostly proven benefits on transfer tasks.



\paragraph{Our approach.}

In this paper, we use a consistency model as a proxy of the flow of a diffusion process to reduce transport costs. 
While not fully solving the alignment issue, we will show that our method present reduced transport costs and better alignment than dedicated OT-based methods.

\section{Consistency Models with Generator-Augmented Flows}
\label{sec:model}
Here, we introduce our method, denoted as generator-augmented flows, which relies on a generator-augmented coupling (GC).
{We capitalize on the true diffusion flow $\mathring{\vf}$ (\textit{i.e.} an ideal consistency model) to map noisy points towards the PF-ODE solution. We present theoretical  and empirical evidences that GC not only reduces the data-noise transport cost but also narrows the gap between consistency distillation and consistency training. We will discuss how to train GC consistency models jointly with $\mathring{\vf}$ in \cref{sec:exp}. }


\subsection{Generator-Augmented Coupling (GC): Definition and Training Loss}
\label{sec:gc}

The solution proposed in this work involves harnessing the diffusion flow, computed from a consistency model, to create a novel form of coupling.
The idea is to leverage the properties and accumulated knowledge within an ideal consistency model, $\mathring{\vf}$, to construct pairs of points. To achieve this, we first sample an intermediate point, which is done as usual by sampling $\rvx_\star \sim p_\star$ and $\rvz \sim p_z$ using the IC between the two distributions, and then predict the data point $\hat{\rvx}_{t_i}$ via the consistency model:
\begin{equation}
    (\rvx_\star, \rvz) \sim q_{\mathrm{I}}; \quad
    \rvx_{t_{i}} =  \rvx_\star + \sigma_{t_i} \rvz; \quad
    \hat{\rvx}_{t_i} = \mathrm{sg}(\mathring{\vf}(\rvx_{t_i}, \sigma_{t_i})).
    \label{eq:intermediate_points_IC}
\end{equation}
Although $\hat{\rvx}_{t_i}$ depends on the timestep $t_i$, it is important to note that it (supposedly) follows the distribution $p_0$.
This $\hat{\rvx}_{t_i}$ is coupled with $\rvz$, thereby defining our \emph{generator-augmented coupling} (GC) $q$, which we use to construct the pair of points $(\tilde{\rvx}_{t_i}, \tilde{\rvx}_{t_{i+1}})$: 
\begin{equation}
    (\hat{\rvx}_{t_i}, \rvz) \sim q; \quad \tilde{\rvx}_{t_{i}} =  \hat{\rvx}_{t_i} + \sigma_{t_{i}} \rvz; \quad \tilde{\rvx}_{t_{i+1}} = \hat{\rvx}_{t_i} + \sigma_{t_{i+1}} \rvz.
    \label{eq:intermediate_points_GI}
\end{equation}
These intermediate points can serve to define a new consistency training loss: 
\begin{multline}
\label{eq:consistency_gc_loss}
    \Ls_{\text{GC}}(\theta) 
    = \E_{q(\hat{\rvx}_{t_i},\rvz), p(\Tilde{\rvx}_{t_i}, \Tilde{\rvx}_{t_{i+1}} | \hat{\rvx}_{t_i}, \rvz)} \\
    \quad \Big[ \lambda(\sigma_{t_i}) \mathcal{D}\Big(\mathrm{sg}(\vf_\theta(\tilde{\rvx}_{t_i},\sigma_{t_i})), \vf_\theta(\tilde{\rvx}_{t_{i+1}} ,\sigma_{t_{i+1}})\Big)\Big].
\end{multline}

\paragraph{Generator-augmented trajectories satisfy the boundary conditions of diffusion processes.}
We note the two following important properties of the distribution of $\tilde{\rvx}_t$:
\begin{align}
    p(\tilde{\rvx}_{0}) = p(\rvx_{0}) \approx p_\star,
    &&
    p(\tilde{\rvx}_{T}) \approx p(\rvx_{T}) \approx p(\sigma_T \rvz).
    \label{eq:boundary_condition_GI}
\end{align}
The first property is achieved thanks to the boundary condition of the consistency model (\textit{c.f.}\ \cref{sec:prelim_diff}) 
, and the second property by construction of the diffusion process which ensures that the noise magnitude is significantly larger than $\hat{\rvx}_{t_i}$ for large $t$. 
However, for the timesteps $t \in (0, T)$ the marginal distributions $p({\rvx}_{t})$ and $p(\tilde{\rvx}_{t})$ do not necessarily coincide.

\subsection{Properties of Generator-Augmented Flows}

Here, we present some properties of generator-augmented flows that motivate them for training consistency models.

\subsubsection{Reducing $\mathcal{R}(\theta)$ with GC}

\begin{figure}
  \centering
  \includegraphics[width=0.95\columnwidth]{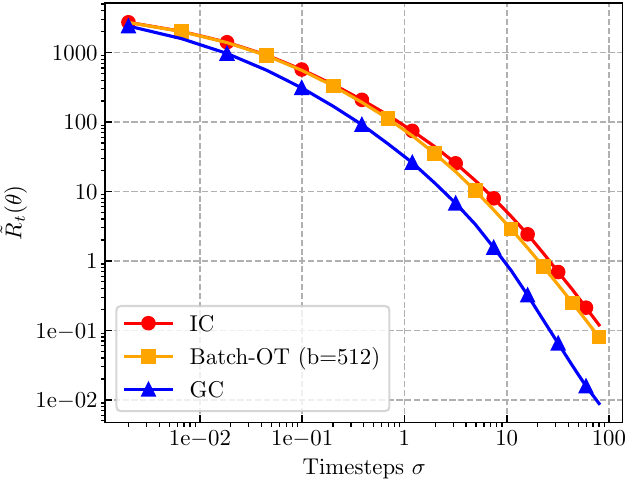}
  \caption{Comparison of $\tilde{\mathcal{R}}_{\text{IC}}$, $\tilde{\mathcal{R}}_{\text{batch-OT}}$, and $\tilde{\mathcal{R}}_{\text{GC}}$ on CIFAR-10. GC exhibits lower values of this quantity for all $\sigma_t$. \label{fig:r_theta}}
\end{figure}

In \cref{theorem_R_theta}, we proved that the continuous-time consistency training objective decomposes into the sum of the consistency distillation objective and a regularizer term: $\Ls_{\text{CT}}(\theta) = \Ls_{\text{CD}}(\theta) + \mathcal{R}(\theta)$. Here, we  study a proxy term for $\mathcal{R(\theta)}$ that is easier to calculate:
\begin{equation}
    \tilde{\mathcal{R}}_t = \E\left[ \left\| \dot{\rvx}_t - \rvv_t(\rvx_t) \right\|^2 \right].
\end{equation}
This quantity measures the expected distance between the true velocity field and its one-sample Monte Carlo estimate. We study $ \tilde{\mathcal{R}}_{t,\mathrm{IC}}$, $\tilde{\mathcal{R}}_{\text{batch-OT}}$, and $ \tilde{\mathcal{R}}_{t,\mathrm{GC}}$. They are the respective proxy regularizer term for each type of probability path.  Note that $\tilde{\mathcal{R}}_{t,\mathrm{GC}}$ depends on the endpoint predictor, a consistency model,  which impacts both probability paths and velocity fields. 
Our goal is to compare those proxy regularizer terms, in order to demonstrate that GC does lead to a smaller discrepancy than IC. We further motivate the use of this proxy, in regards with \cref{theorem_R_theta}, in \cref{subsec:proxy}.

In the following theorem, proved in \cref{app:proof_proxy}, we show that $\tilde{\mathcal{R}}_t$ decays faster for GC than for IC. 
%


\begin{theorem}\label{theorem_R_IC_GC}
    Assume that the data distribution contains more than a single point. Also, assume that the generator-augmented coupling between the predicted data point $\hat{\rvx}_t$ and noise $\rvz$ is computed via an ideal consistency model $\mathring{\vf}$, \textit{i.e.}, the flow of the PF-ODE. Then, as $t \to \infty$,
    \begin{equation}
        \tilde{\mathcal{R}}_{t,\mathrm{GC}} \ll \tilde{\mathcal{R}}_{t,\mathrm{IC}}. 
    \end{equation}
\end{theorem}


\paragraph{Empirical validation. } Evaluating $\tilde{\mathcal{R}}_{t}$ requires computing the difference between the sample path derivative $\dot{\rvx}_t$ and the velocity field $\rvv_t(\rvx_t)$.
In the EDM setting, this difference can be approximated using a denoiser. Indeed, $\dot{\rvx}_t = \rvz$ and $\rvv_t(\rvx_t) = \E[\dot{\rvx}_t | \rvx_t] = \E[\rvz | \rvx_t] = \E[\frac{\rvx_t -\rvx_\star}{t} | \rvx_t] = \frac{1}{t}(\rvx_t - \boldsymbol{D}_\star(\rvx_t, t))$ with an optimal denoiser $\boldsymbol{D}_\star$. The optimal denoiser can be approximated by a denoiser network $\boldsymbol{D}_\phi$. Finally, we have: $\dot{\rvx}_t - \rvv_t(\rvx_t) \approx \rvz - \frac{1}{t}(\rvx_t - \boldsymbol{D}_\phi(\rvx_t,t))$.
Since IC, batch-OT, and GC define different $p_t$'s and $\rvv_t$'s, we train a different denoiser $\boldsymbol{D}_\phi$ for each coupling.   
In \cref{fig:r_theta}, we report the results from the comparison of the three proxy terms on CIFAR-10. 
We observe that $\tilde{\mathcal{R}}_{t,\text{GC}} < \tilde{\mathcal{R}}_{t,\text{batch-OT}} <  \tilde{\mathcal{R}}_{t,\text{IC}}$ and that the gap increases with $t$, corroborating our theoretical findings (\cref{theorem_R_IC_GC}).

\subsubsection{Reducing Transport Cost with GC}

\begin{figure}
    \centering
    \includegraphics[width=0.95\columnwidth]{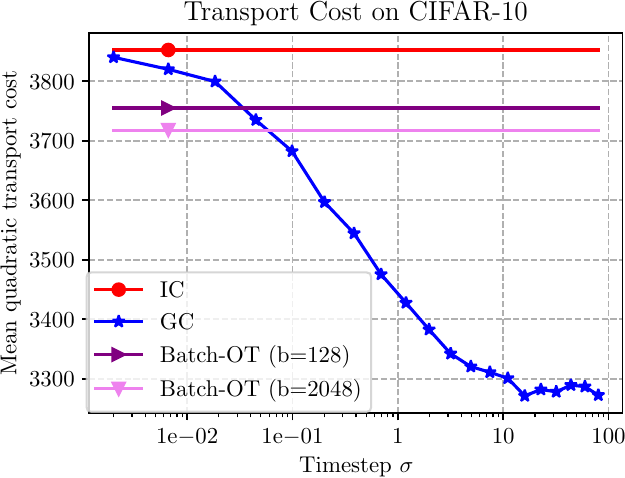}
    \caption{
       Comparison of transport costs between IC, batch-OT, and GC on CIFAR-10. \label{fig:transport_cost}
    }
\end{figure}

Here, we investigate the average transport cost between the noise $\rvz \sim p_z$ and the predicted data point $\hat{\rvx} \sim p_\star$ as a measure of the efficiency of the data-noise coupling. Recall that the diffusion process is given by $\rvx_t = \rvx_\star + \sigma_t \rvz$. Then, knowing that the consistency model $\mathring{\vf}$ satisfying the boundary condition $\mathring{\vf}(\rvx_0, \sigma_0) = \rvx_0$, we define the function $c(t)$ as:
\begin{equation}
    c(t) = \E_{q_\mathrm{I}(\rvx_\star,\rvz)} \left[ \left\| \mathring{\vf}(\rvx_t, \sigma_t) - \rvz \right\|^2 \right].
    \label{eq:transportation_cost}
\end{equation}
$c(0) = \E_{q_\mathrm{I}(\rvx_\star,\rvz)} [ \| \rvx_0 - \rvz \|^2 ]$ and $c(t)$ represent the transport costs of, respectively, IC and GC. We show below, with proofs in \cref{app:proof_transport}, that $c(t)$ is decreasing for $\sigma_t$ close to zero and for large $\sigma_t$.

\begin{lemma}[\textbf{Transport cost of GC coupling}]\label{lemma:transport_cost}
    Assume that $\mathring{\vf}$ is a continuously differentiable function representing the ground-truth consistency model, \textit{i.e.}\ the flow of the PF-ODE induced by the diffusion process $\rvx_t$. Define $\rvw_t = \rvz - \E[\rvz | \rvx_t] = \frac{1}{\dot{\sigma}_t} (\dot{\rvx}_t - \E[\dot{\rvx}_t \mid \rvx_t])$. Then:
    \begin{align}
        c'(t) &= - 2 \dot{\sigma}_t \E\left[ \left< \frac{\partial \mathring{\vf}}{\partial \rvx} (\rvx_t, \sigma_t) \cdot \rvw_t, \rvw_t \right> \right].
    \end{align}
\end{lemma}

\begin{corollary}[\textbf{Decreasing transport cost of GC coupling in $t\to 0^+$}] \label{cor:transport_t0}
    There exists a $t_* > 0$ such that for all $t \in [0,t_*]$, the derivative of $c(t)$ takes the form $  c'(t) = - 2 \dot{\sigma}_t a_t$ with $a_t > 0$. Hence for $\dot{\sigma}_t$  positive, the cost is decreasing.  In particular, in the EDM setting where $\sigma_t = t$, $c(t)$ is decreasing for small $t$.
\end{corollary}
The proof of this corollary proceeds by noting that for $t=0$, the consistency model $\mathring{\vf}(\rvx,t)$ is an identity function, its Jacobian is an identity matrix, and thus $a_t = \E [ \|\rvw_t\| ^2]$. Using the continuity of Jacobian elements and invoking intermediate value theorem on $a_t$ concludes the proof.

\begin{corollary}[\textbf{Decreasing transport cost of GC coupling in $t \approx t_{\max}$}] \label{cor:transport_tmax}
    Assume that the consistency model $\mathring{\vf}(x, \sigma)$ is a scaling function $\mathring{\vf}(\rvx, \sigma_t) = \frac{\sigma_0}{\sigma_t} \rvx$. Then, we have $ c'(t) = - \frac{ 2 \dot{\sigma}_t \sigma_0}{ \sigma_t } \E[\| \rvw_t \|^2]$. In particular, $c(t)$ is decreasing whenever $\sigma_t$ is increasing.
\end{corollary}
We note that, while the assumption of the consistency model being a scaling function is strong, it nonetheless bears some degree of truth for $t \approx t_{\max}$, see \cref{lemma:cm_almost_scaling} of \cref{sec:proofs}.

\paragraph{Experimental validation.}
As stressed in \cref{sec:coupling}, a line of work has brought evidence that reducing the transport cost between noise and data distributions could fasten the training and help produce better samples. We compare the quadratic transport costs involved in IC, batch-OT \citep{pooladian2023multisample, dou2024a}, and GC (resp. $c(0)$, $c_{\text{OT}}(0)$, and $c(t)$). Results are presented in \cref{fig:transport_cost}. Interestingly, GC reduces transport cost more than batch-OT on CIFAR-10 because batch-OT is tied to the batch data points $\rvx_t$ whereas our computed $\hat{\rvx}_t$ are not.

\section{Training With Generator-Augmented Flows for Image Generation \label{sec:exp}}

{In this section, we present a methodology to train consistency models with GC on unconditional image generation. 
To construct points drawn  from GC trajectories ($\tilde{\rvx}_{t_{i}}$), our theory requires an optimal predictor $\mathring{\vf}$ on intermediate points drawn from IC ($\rvx_{t_{i}}$). 
Thus, this lets us two potential training strategies: \textit{(i)} pre-train an IC generator, and leverage it to construct GC trajectories that train a GC model; \textit{(ii)} a joint learning strategy: train a single consistency model from scratch with both types of trajectories. Note that in this second setting, the model is unique: $\mathring{\vf} = \vf_\theta $. 
The second option is more appealing, since it is a simple one-stage training. We demonstrate that the joint learning approach improves performance and accelerates convergence compared to standard consistency models. }

Our experiments are done on the following datasets: CIFAR-10 \citep{cifar}, ImageNet \citep{imagenet}, CelebA \citep{Liu2015} and LSUN Church \citep{yu2015lsun}.
For the evaluation metrics, we report the Fréchet Inception Distance (FID, \citet{heusel2017gans}), Kernel Inception Distance (KID, \citet{binkowski2018demystifying}), and Inception Score (IS, \citet{salimans2016improved}). Most of our experiments are based on the improved training techniques for consistency models from \citet{song2023improved}, denoted iCT-IC. Moreover, we present some results in the setting of Easy Consistency Tuning (ECT, \citet{ect}).  
Details are provided in \cref{app:details}. The code is shared in the supplementary material and will be open-sourced upon publication for reproducibility.

\subsection{GC with Pre-Trained Endpoint Predictor}

\begin{figure}
    \centering
    \includegraphics[width=0.95\columnwidth]{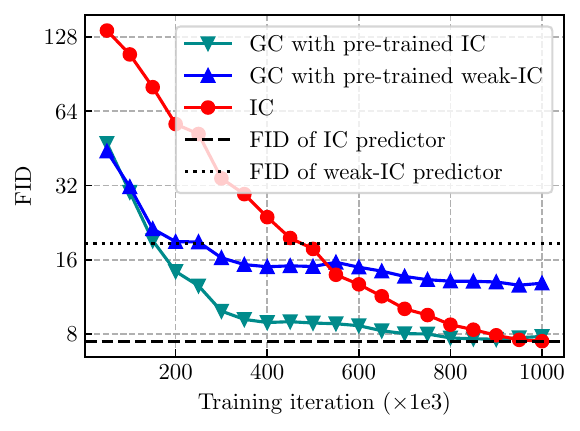}
    \caption{Performance of GC w.r.t.\ the performance of the predictor on CIFAR-10.}
    \label{fig:GC_pretrained_IC}
\end{figure}

{Our theoretical results assume having access to an ideal generator on IC trajectories, meaning that the generator approximates the diffusion flow output.} To train a consistency model on GC, we can thus rely on a separate endpoint predictor pre-trained on IC (iCT-IC): $\mathring{\vf} \equiv \vg_\phi$ (cf. \cref{sec:gc}). This network predicts the endpoint: $\hat{\rvx}_{t_i} = \vg_\phi(\rvx_{t_i}, \sigma_{t_i})$.
During the training of the consistency model on GC, $\vg_\phi$ is kept frozen and considered a proxy of the true flow, as in our theoretical results.
In \cref{fig:GC_pretrained_IC}, we report the performance of consistency models on CIFAR-10 trained with GC using two different $\vg_\phi$: \emph{(i)} a $\vg_\phi$ fully trained as standard iCT-IC with 100k training steps; \emph{(ii)} a weak $\vg_\phi$ partially trained as iCT-IC with 20k training steps.

\begin{observation} 
Using a partially pre-trained and frozen endpoint predictor, trained on IC trajectories, allows to train a consistency model with GC and which converges faster. However, the performance of the GC model depends on the quality of the endpoint predictor on IC trajectories.
\end{observation}

It is important to note that this setup is not practical, as it requires pre-training a standard consistency model. We aim for a training methodology that accelerates convergence and improves performance when training from scratch, without doubling the number of required training iterations.

\begin{table*}
\caption{iCT-IC is the standard improved consistency model with independent coupling \citep{song2023improved}; iCT-OT is iCT with minibatch optimal transport coupling \citep{pooladian2023multisample, dou2024a}; iCT-GC ($\mu=0.5$) is our proposed GC with joint learning. 
\label{tab:full_results}
}
\centering
\begin{tabular}{llccc} 
\toprule
Dataset & Model & FID $\downarrow$ & KID ($\times 10^{2}$) $\downarrow$ & 
IS $\uparrow$ \\ 
\midrule
\multirow{3}{*}{CIFAR-10} & iCT-IC & 7.42 $\pm$ 0.04 & 0.44 $\pm$ 0.03 & 8.76 $\pm$ 0.06  \\
& iCT-OT & 6.75 $\pm$ 0.04 & 0.36 $\pm$ 0.04 & 8.86 $\pm$ 0.09  \\
& iCT-GC ($\mu=0.5$) & \textbf{5.95} $\pm$ 0.05 & \textbf{0.26} $\pm$ 0.02 & \textbf{9.10} $\pm$ 0.05  \\
\midrule
\multirow{3}{*}{ImageNet ($32\times32$)} & iCT-IC & 14.89 $\pm$ 0.17 & 1.23 $\pm$ 0.05 & 9.46 $\pm$ 0.06 \\
& iCT-OT & 14.13 $\pm$ 0.17 & 1.18 $\pm$ 0.05 & 9.62 $\pm$ 0.06 \\
& iCT-GC ($\mu=0.5$) & \textbf{13.99} $\pm$ 0.28 & \textbf{1.13} $\pm$ 0.03 & \textbf{9.77} $\pm$ 0.07 \\
\midrule
\multirow{3}{*}{CelebA ($64\times64$)} & iCT-IC & 15.82 $\pm$ 0.13 & 1.31 $\pm$ 0.04 & 2.33 $\pm$ 0.00  \\
& iCT-OT & 13.63 $\pm$ 0.13 & 1.09 $\pm$ 0.03 & 2.40 $\pm$ 0.01  \\
& iCT-GC ($\mu=0.5$) & \textbf{11.74} $\pm$ 0.08 & \textbf{0.91} $\pm$ 0.04 & \textbf{2.45} $\pm$ 0.01  \\
\midrule
\multirow{3}{*}{LSUN Church ($64\times64$)} & iCT-IC & 10.58 $\pm$ 0.11 & 0.73 $\pm$ 0.03 & 1.99 $\pm$ 0.01 \\
& iCT-OT & \textbf{9.71} $\pm$ 0.13 & \textbf{0.64} $\pm$ 0.03 & 2.00 $\pm$ 0.01 \\
& iCT-GC ($\mu=0.5$) & 9.88 $\pm$ 0.07 & 0.66 $\pm$ 0.04 & \textbf{2.14} $\pm$ 0.01  \\
\bottomrule
\end{tabular}
\end{table*}

\subsection{GC from Scratch with Joint Learning}
\begin{figure}
    \centering
    \includegraphics[width=0.95\columnwidth]{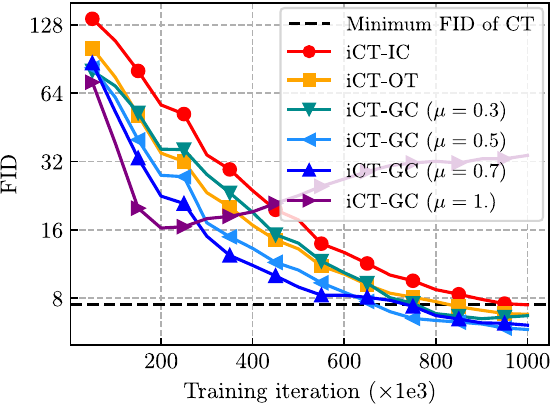}
    \caption{Consistency models trained with GC with joint learning converges faster and outperforms consistency models trained with IC or minibatch-OT on CIFAR-10.}
    \label{fig:mixing}
\end{figure}

In this section, we propose to learn simultaneously a single model on IC and GC trajectories from the start of the training, \textit{i.e.}\ $\mathring{\vf} \equiv \mathrm{sg}(\vf_\theta)$ (cf. \cref{sec:gc}). Thereby, we combine the training of the ideal IC predictor with the training of GC model based on this predictor.
We introduce a joint learning factor $\mu$: at each training step, training pairs are drawn from GC with probability $\mu$, while the remaining pairs are drawn from standard IC. The loss can be written on average as: 
\begin{equation}
    \Ls_{\text{GC-}\mu}(\theta) = \mu \Ls_{\text{GC}}(\theta) + (1-\mu) \Ls_{\text{CT}}(\theta)
\end{equation}

We denote this joint learning procedure as GC ($\mu=\cdot$). Hence, GC ($\mu=0$) corresponds to the standard IC procedure, while GC ($\mu=1$) corresponds to training only with GC points.
Note that GC ($\mu=1$) is not expected to work, since our theoretical guarantees assume an optimal IC predictor. The detailed algorithm is presented in \cref{alg:consistency_model_w_GI} in Appendix. We apply this joint learning to four image datasets, and include comparisons to iCT with batch-OT \citep{dou2024a} as an additional baseline. Results across multiple datasets and metrics are presented in \cref{tab:full_results}, and visual examples are shown in \cref{fig:celeba_generated_img} in Appendix.

\begin{observation} Joint learning of IC and GC trajectories consistently improves results compared to the base IC model and outperforms batch-OT in most cases. \end{observation}

As shown in \cref{fig:mixing}, we observe an interesting interpolation phenomenon between $\mu=0$ and $\mu=1$. For $\mu=0$, we recover the steady FID improvement typical of IC training. As $\mu$ increases, the convergence of the generative model accelerates.
For $0.3 \leq \mu \leq 0.7$, on CIFAR-10, convergence speed and final FID are improved compared to IC and batch-OT models.
For $\mu=1$, the FID score decreases faster than other configurations early in the training process, but it soons increases as training progresses further. 
It is explained by the poor performance of the predictions on IC yielding a deviation from the ideal IC predictor from \cref{sec:model}.  For the other datasets, we simply chose $\mu=0.5$ and report those results. 
We provide further detail on the sensitivity of our results to the choice of $\mu$ in \cref{sec:ablation,app:details}.

\subsection{GC in the ECT Setting}

\begin{table}
    \caption{Performance of IC and GC consistency models trained in the ECT setting \citep{ect}. Short training: $4k$ iterations. Long training: $100k$ iterations.}\label{tab:ect}
 \vspace{\baselineskip}
    \centering
    \begin{tabular}{l r} 
    \toprule  
     {Model} &  {FID} $\downarrow$ \\ 
    \midrule
     \multicolumn{2}{l}{\textit{CIFAR-10 (Short Training)}} \\
     \quad ECT-IC &  7.37 $\pm$ 0.05 \\  
     \quad ECT-GC ($\mu=0.3$) & \textbf{6.41} $\pm$ 0.05 \\ 
    \cmidrule(lr){1-2}
     \multicolumn{2}{l}{\textit{CIFAR-10 (Long Training)}} \\
     \quad ECT-IC & 4.11 $\pm$ 0.03 \\
     \quad ECT-GC ($\mu=0.3$) & \textbf{3.74} $\pm$ 0.04 \\
    \cmidrule(lr){1-2}
     \multicolumn{2}{l}{\textit{FFHQ $64\times64$ (Short Training)}} \\
     \quad ECT-IC & 13.29 $\pm$ 0.10 \\
     \quad ECT-GC ($\mu=0.3$) & \textbf{11.73} $\pm$ 0.09 \\
    \cmidrule(lr){1-2}
     \multicolumn{2}{l}{\textit{FFHQ $64\times64$ (Long Training)}} \\
     \quad ECT-IC & 9.68 $\pm$ 0.06 \\
     \quad ECT-GC ($\mu=0.3$) & \textbf{8.51} $\pm$ 0.09 \\
    \cmidrule(lr){1-2}
     \multicolumn{2}{l}{\textit{ImageNet $64\times64$ Cond. (Short Training)}} \\
     \quad ECT-IC & 10.82 $\pm$ 0.18 \\
     \quad ECT-GC ($\mu=0.3$) & \textbf{10.31} $\pm$ 0.22 \\
    \cmidrule(lr){1-2}
     \multicolumn{2}{l}{\textit{ImageNet $64\times64$ Cond. (Long Training)}} \\
     \quad ECT-IC & \textbf{5.84} $\pm$ 0.21 \\
     \quad ECT-GC ($\mu=0.3$) & 6.39 $\pm$ 0.20 \\
    \bottomrule
    \end{tabular}
\end{table}

As an additional experiment, we explore the recent ECT setting \citep{ect} on CIFAR-10, where consistency models are fine-tuned from a pre-trained diffusion model. This approach enables training high-quality consistency models in one GPU-hour, though it requires an already trained diffusion model.

We compare IC and GC trajectories in this setting, with both short (approximately one GPU-hour) and long (100k steps, 1 GPU-day) training times. Using the referenced hyper-parameters selected by \citet{ect}, we observe a consistent advantage for GC, with an optimal $\mu$ value of $0.3$. These preliminary results, summarized in \cref{tab:ect}, align with our previous findings on the iCT setting, further supporting the effectiveness of GC.

\section{Conclusion}
In this paper, we identify a discrepancy between consistency training and consistency distillation. Building on this theoretical analysis, we introduce generator-augmented flows and show that they reduce a proxy term measuring this discrepancy. Additionally, generator-augmented flows decrease the data-to-noise transport cost, as demonstrated by theory and experiments. Finally, we derive practical algorithms for training consistency models using generator-augmented flows and demonstrate improved empirical performance.


\section*{Impact Statement}

If used in large-scale generative models, notably in text-to-image models, this work may increase potential negative impacts of deep generative models such as \textit{deepfakes} \citep{fallis2020epistemic}. 

\paragraph{Acknowledgements.} This research was funded by grant Nos. 2023-00262155, 2024-00460980 and 2025-02304717 (IITP) funded by the Korea government (the Ministry of Science and ICT).

\bibliography{refs}
\newpage
\appendix
\onecolumn

\section{Proofs}
\label{sec:proofs}


\subsection{Continuous-Time Consistency Objectives}
\label{app:proof_discrepancy}

\begin{reptheorem}{theorem_R_theta}[\textbf{Discrepancy between consistency distillation and consistency training objectives}]
    Assume that the distance function is given by $\mathcal{D}(\rvx, \mathbf{y}) = \varphi(\| \rvx - \mathbf{y}\|)$ for a continuous convex function $\varphi : [0, \infty) \to [0, \infty)$ with $\varphi(x) \sim Cx^{\alpha}$ as $x \to 0^+$ for some $C > 0$ and $\alpha \geq 1$, and that the timesteps are equally spaced, i.e., $t_i = \frac{i T}{N}$. Furthermore, assume that the Jacobian $\frac{\partial \vf_\theta}{\partial \rvx}$ does not vanish identically. Then the following assertions hold:
    \begin{enumerate}
        \item[(i)] The scaled consistency losses $N^\alpha \Ls_\mathrm{CD}(\theta)$ and $N^\alpha \Ls_\mathrm{CT}(\theta)$ converge as $N \to \infty$. Moreover, the minimization objectives corresponding to these limiting scaled consistency losses are not equivalent, and their difference is given by:
        \begin{equation}
            \lim_{N\to\infty} N^\alpha \left[ \Ls_\mathrm{CT}(\theta) - \Ls_\mathrm{CD}(\theta) \right]
            = CT^{\alpha-1} \mathcal{R}(\theta),
        \end{equation}
        where $\mathcal{R}(\theta)$ is defined by
        \begin{equation}
            \mathcal{R}(\theta)
            = \int_{0}^{T} \lambda(\sigma_t) \E \left[ \left\| \partial_\mathrm{CT}\vf_\theta \right\|^\alpha - \left\| \partial_\mathrm{CD}\vf_\theta \right\|^\alpha \right] \, \dif t
        \end{equation}
        and satisfies $\mathcal{R}(\theta) > 0$, with
        \begin{equation}
            \partial_\mathrm{CT}\vf_\theta
            = \frac{\partial \vf_\theta}{\partial \sigma}(\rvx_t, \sigma_t) \dot{\sigma}_t + \frac{\partial \vf_\theta}{\partial \rvx}(\rvx_t, \sigma_t) \cdot \dot{\rvx}_t,
        \end{equation}
        \begin{equation}
            \partial_\mathrm{CD}\vf_\theta
            = \frac{\partial \vf_\theta}{\partial \sigma}(\rvx_t, \sigma_t) \dot{\sigma}_t + \frac{\partial \vf_\theta}{\partial \rvx}(\rvx_t, \sigma_t) \cdot \rvv_t(\rvx_t).
        \end{equation}
        
        In particular, if $\alpha=2$,
        \begin{equation}
            \mathcal{R}(\theta)
            = \int_{0}^{T} \lambda(\sigma_t) \E \left[ \left\| \frac{\partial \vf_\theta}{\partial \rvx}(\rvx_t, \sigma_t) \cdot \left( \dot{\rvx}_t - \rvv_t(\rvx_t) \right) \right\|^2 \right] \, \dif t.
        \end{equation}

        \item[(ii)] The scaled gradient $N^{\alpha-1} \nabla_\theta \Ls_\mathrm{CD}(\theta)$ and $N^{\alpha-1} \nabla_\theta \Ls_\mathrm{CT}(\theta)$ converge as $N \to \infty$. Moreover, if $\alpha \neq 2$, then their respective limits are not identical as functions of $\theta$:
        \begin{equation}
            \lim_{N\to\infty} N^{\alpha-1} \nabla_\theta \Ls_\mathrm{CT}(\theta)
            \neq \lim_{N\to\infty} N^{\alpha-1} \nabla_\theta \Ls_\mathrm{CD}(\theta).
        \end{equation}
    \end{enumerate}
\end{reptheorem}

\begin{proof}[Proof.]
    \textit{(i)} Note that $\partial_\mathrm{CD}\vf_\theta$ and $\partial_\mathrm{CT}\vf_\theta $ satisfy:
    \begin{align}
        \partial_\mathrm{CT}\vf_\theta(\rvx_t, \sigma_t)
        &= \frac{\partial}{\partial t} \vf_\theta(\rvx_t, \sigma_t),
        &
        \partial_\mathrm{CD}\vf_\theta(\rvx_t, \sigma_t)
        &= \E\left[ \frac{\partial}{\partial t} \vf_\theta(\rvx_t, \sigma_t) \middle| \rvx_t \right].
        \label{eq:material_derivatives}
    \end{align}
    Here, the second equality follows by noting that $\rvv_t(\rvx_t) = \E[\dot{\rvx}_t | \rvx_t]$ and all the other terms in the expansion of $\frac{\partial}{\partial t} \vf_\theta(\rvx_t, \sigma_t)$ are completely determined once the value of $\rvx_t$ is known.
    
    Now, we use Taylor's theorem to expand the difference between $\vf_\theta(\rvx_{t_{i+1}},\sigma_{t_{i+1}})$ and $\vf_\theta(\rvx_{t_{i}}^{\Phi}, \sigma_{t_i})$ in the consistency distillation loss, \cref{eq:consistency_distill_loss}. Together with the definition of $\rvx_{t_{i}}^{\Phi}$, \cref{eq:score_update}, this yields:
    \begin{align}
        &\vf_\theta(\rvx_{t_{i+1}},\sigma_{t_{i+1}})
        - \vf_\theta(\rvx_{t_{i}}^{\Phi}, \sigma_{t_i}) \nonumber \\
        &=
        \frac{\partial \vf_\theta}{\partial \sigma}(\rvx_{t_{i+1}}, \sigma_{t_{i+1}}) \cdot ( \sigma_{t_{i+1}} - \sigma_{t_i} )
        + \frac{\partial \vf_\theta}{\partial \rvx}(\rvx_{t_{i+1}}, \sigma_{t_{i+1}}) \cdot ( \rvx_{t_{i+1}} - \rvx_{t_i}^{\Phi} )
        + o(t_{i+1} - t_i) \\
        &=
        \partial_\mathrm{CD}\vf_\theta (\rvx_{t_{i+1}}, \sigma_{t_{i+1}}) \cdot (t_{i+1} - t_i)
        + o(t_{i+1} - t_i).
    \end{align}
    Similarly, by expanding the difference between $\vf_\theta(\rvx_{t_{i+1}},\sigma_{t_{i+1}})$ and $\vf_\theta(\rvx_{t_{i}},\sigma_{t_{i}})$ in \cref{eq:consistency_training_loss},
    \begin{align}
        &\vf_\theta(\rvx_{t_{i+1}},\sigma_{t_{i+1}})
        - \vf_\theta(\rvx_{t_{i}}, \sigma_{t_i}) \nonumber \\
        &=
        \frac{\partial \vf_\theta}{\partial \sigma}(\rvx_{t_{i+1}}, \sigma_{t_{i+1}}) \cdot ( \sigma_{t_{i+1}} - \sigma_{t_i} )
        + \frac{\partial \vf_\theta}{\partial \rvx}(\rvx_{t_{i+1}}, \sigma_{t_{i+1}}) \cdot ( \rvx_{t_{i+1}} - \rvx_{t_i} )
        + o(t_{i+1} - t_i) \\
        &=
        \partial_\mathrm{CT}\vf_\theta (\rvx_{t_{i+1}}, \sigma_{t_{i+1}}) \cdot (t_{i+1} - t_i)
        + o(t_{i+1} - t_i).
    \end{align}
    Therefore, for each $\bullet \in \{\mathrm{CD}, \mathrm{CT}\}$,
    \begin{align}
        N^\alpha \Ls_{\bullet}(\theta)
        &= N^{\alpha} \cdot \frac{1}{N} \sum_{i=0}^{N-1} \lambda(\sigma_{t_i}) \E \left[ C \left\| \partial_\bullet \vf_\theta (\rvx_{t_{i+1}}, \sigma_{t_{i+1}}) \right\|^{\alpha}(1 + o(1)) \right] \cdot (t_{i+1} - t_i)^{\alpha} \\
        &= C T^{\alpha-1} \sum_{i=0}^{N-1} \lambda(\sigma_{t_i}) \E \left[ \left\| \partial_\bullet \vf_\theta (\rvx_{t_{i+1}}, \sigma_{t_{i+1}}) \right\|^{\alpha}(1 + o(1)) \right] \cdot (t_{i+1} - t_i) \\
        &\to C T^{\alpha-1} \int_{0}^{T} \lambda(\sigma_t) \E \left[ \left\| \partial_\bullet \vf_\theta (\rvx_t, \sigma_t) \right\|^{\alpha} \right] \, \dif t \label{eq:limit_loss}
    \end{align}
    in the continuous-time limit as $N\to\infty$.
    
    For simplicity of notation, we write
    \begin{align}
        \Ls^{\infty}_{\bullet}(\theta) = \lim_{N\to\infty} N^\alpha \Ls_{\bullet}(\theta)
    \end{align}
    for each $\bullet \in \{\mathrm{CD}, \mathrm{CT}\}$. Then, from the formula for the limiting losses $\Ls_\bullet^\infty(\theta)$, \cref{eq:limit_loss}, we immediately obtain
    \begin{equation}
        \Ls_\mathrm{CT}^\infty(\theta) - \Ls_\mathrm{CD}^\infty(\theta)
        = C T^{\alpha-1} \int_{0}^{T} \lambda(\sigma_t) \E \left[ \left\| \partial_\mathrm{CT} \vf_\theta (\rvx_t, \sigma_t) \right\|^{\alpha} - \left\| \partial_\mathrm{CD} \vf_\theta (\rvx_t, \sigma_t) \right\|^{\alpha} \right] \, \dif t.
    \end{equation}
    Now, we specialize in the case $\alpha = 2$ and invoke the general observation that, for any random vectors $\rvx$ and $\rvy$, the following identity holds:
    \begin{align}
        \E\left[\|\rvx\|^2 - \|\E[\rvx | \rvy]\|^2\right]
        &= \E\left[\|\rvx - \E[\rvx | \rvy]\|^2\right].
    \end{align}
    This can be easily proved by expanding the squared Euclidean norm as the inner product and applying the law of iterated expectations. Plugging in $\rvx \leftarrow \frac{\partial}{\partial t} \vf_\theta(\rvx_t, \sigma_t)$ and $\rvy \leftarrow \rvx_t$, and noting that $\partial_\mathrm{CD} \vf_\theta (\rvx_t, \sigma_t) = \E\left[ \partial_\mathrm{CT} \vf_\theta (\rvx_t, \sigma_t) \mid \rvx_t\right]$ by \cref{eq:material_derivatives}, it follows that
    \begin{align}
        \Ls_\mathrm{CT}^\infty(\theta) - \Ls_\mathrm{CD}^\infty(\theta)
        &= C T \int_{0}^{T} \lambda(\sigma_t) \E \left[ \left\| \partial_\mathrm{CT} \vf_\theta (\rvx_t, \sigma_t) - \partial_\mathrm{CD} \vf_\theta (\rvx_t, \sigma_t) \right\|^{2} \right] \, \dif t \\
        &= C T \int_{0}^{T} \lambda(\sigma_t) \E \left[ \left\| \frac{\partial \vf_\theta}{\partial \rvx}(\rvx_t, \sigma_t) \cdot \left( \dot{\rvx}_t - \rvv_t(\rvx_t) \right) \right\|^{2} \right] \, \dif t.
    \end{align}
    
    Next, we establish the positivity of $\mathcal{R}(\theta)$. To this end, note that $\|\cdot\|^\alpha$ is a convex function for $\alpha \geq 1$. By invoking the conditional Jensen's inequality, we find that the expectation inside the limiting scaled consistency training losses, \cref{eq:limit_loss} satisfy:
    \begin{align}
        \E \Bigl[ \left\| \partial_\text{CT} \vf_\theta (\rvx_t, \sigma_t) \right\|^{\alpha} \Bigr]
        &= \E \Biggl[ \biggl\| \frac{\partial}{\partial t} \vf_\theta(\rvx_t, \sigma_t) \biggr\|^{\alpha} \Biggr]
        = \E \Biggl[ \E \biggl[ \biggl\| \frac{\partial}{\partial t} \vf_\theta(\rvx_t, \sigma_t) \biggr\|^{\alpha} \biggm| \rvx_t \biggr] \Biggr] \\
        &\geq \E \Biggl[ \biggl\| \E \biggl[ \frac{\partial}{\partial t} \vf_\theta(\rvx_t, \sigma_t) \biggm| \rvx_t \biggr] \biggr\|^{\alpha} \Biggr]
        = \E \Bigl[ \left\| \partial_\text{CD} \vf_\theta (\rvx_t, \sigma_t) \right\|^{\alpha} \Bigr].
    \end{align}
    Integrating both sides with respect to $\lambda(\sigma_t) \, \dif t$, we obtain the desired inequality. The Jensen's inequality also tells that the equality holds precisely when $\frac{\partial}{\partial t} \vf_\theta(\rvx_t, \sigma_t) = \E [ \frac{\partial}{\partial t} \vf_\theta(\rvx_t, \sigma_t) | \rvx_t ]$ holds, or equivalently, $ \frac{\partial \vf_\theta}{\partial \rvx}(\rvx_t, \sigma_t) \cdot (\dot{\rvx_t} - \E[\dot{\rvx}_t | \rvx_t]) = 0$. However, given the value of $\rvx_t$, the quantity $\dot{\rvx}_t$ can assume an arbitrary value in $\mathds{R}^d$ because the conditional density of $\dot{\rvx}_t = \dot{\sigma}_t \rvz$ given $\rvx_t$ is strictly positive everywhere. Consequently, the equality condition implies $\frac{\partial \vf_\theta}{\partial \rvx} = 0$. Since this contradicts the assumption of the theorem, the strict inequality between the two limiting losses must hold.
    
    Finally, recall that the continuous-time consistency distillation loss, $\Ls^{\infty}_\mathrm{CD}(\theta)$, is given by
    \begin{align}
        \Ls^{\infty}_\mathrm{CD}(\theta)
        = CT^{\alpha-1} \int_{0}^{T} \lambda(\sigma_t) \E \left[ \left\| \frac{\partial \vf_\theta}{\partial \sigma}(\rvx_t, \sigma_t) \dot{\sigma}_t + \frac{\partial \vf_\theta}{\partial \rvx}(\rvx_t, \sigma_t) \cdot \rvv_t(\rvx_t) \right\|^{\alpha} \right] \, \dif t.
    \end{align}
    Similarly, the continuous-time consistency training loss, $\Ls^{\infty}_\mathrm{CT}(\theta)$, is given by
    \begin{align}
        \Ls^{\infty}_\mathrm{CT}(\theta)
        = CT^{\alpha-1} \int_{0}^{T} \lambda(\sigma_t) \E \left[ \left\| \frac{\partial \vf_\theta}{\partial \sigma}(\rvx_t, \sigma_t) \dot{\sigma}_t + \frac{\partial \vf_\theta}{\partial \rvx}(\rvx_t, \sigma_t) \cdot \dot{\rvx}_t \right\|^{\alpha} \right] \, \dif t.
    \end{align}
    Since $\rvv_t(\rvx_t) = \E [ \dot{\rvx}_t | \rvx_t]$ and $\E[\|\dot{\rvx}_t - \E[\dot{\rvx}_t]\|^2] > \E[\|\rvv_t(\rvx_t) - \E[\dot{\rvx}_t] \|^2]$, it follows that $\Ls^{\infty}_\mathrm{CT}(\theta)$ penalizes the Jacobian $\frac{\partial \vf_\theta}{\partial \rvx}$ more strongly than $\Ls^{\infty}_\mathrm{CD}(\theta)$ does. Therefore, the two limiting consistency losses do not define equivalent objectives.

    \textit{(ii)}
    Using the convexity of $\varphi$, we can show that $\varphi'(x) \sim C \alpha x^{\alpha -1}$ as $x \to 0^+$. Combining this with the vector calculus formula $\nabla_\rvy \|\rvy\| = \frac{\rvy}{\| \rvy \|}$, we get $\nabla_\rvy \varphi(\| \rvy \|) \approx C\alpha \|\rvy\|^{\alpha-2}\rvy$ for small $\rvy$. From this, we can estimate the gradient of the distance between $\mathrm{sg}\bigl(\vf_\theta(\rvx_{t_{i}}^{\Phi}$, $\sigma_{t_i})\big)$ and $\vf_\theta(\rvx_{t_{i+1}}, \sigma_{t_{i+1}})$ with respect to the model parameter $\theta$ as:
    \begin{align}
        &\nabla_\theta \mathcal{D}\bigl(\mathrm{sg}\bigl(\vf_\theta(\rvx_{t_{i}}^{\Phi}, \sigma_{t_i})\big),\vf_\theta(\rvx_{t_{i+1}},\sigma_{t_{i+1}})\bigr)\nonumber \\
        &=  (1 + o(1)) C\alpha
        \left[ \| \partial_\mathrm{CD}\vf_\theta \|^{\alpha - 2} (\partial_\mathrm{CD}\vf_\theta)^\top \frac{\partial \vf_\theta}{\partial \theta} \right] \cdot (t_{i+1} - t_i)^{\alpha-1}
    \end{align}
    Here, the expression $\| \partial_\mathrm{CD}\vf_\theta \|^{\alpha - 2} (\partial_\mathrm{CD}\vf_\theta)^\top \frac{\partial \vf_\theta}{\partial \theta}$ in the square bracket is evaluated at $(\rvx_{t_{i+1}}, \sigma_{t_{i+1}})$. Similarly, the gradient of the distance between $\vf_\theta(\rvx_{t_{i}}, \sigma_{t_{i}})$ and $\vf_\theta(\rvx_{t_{i+1}}, \sigma_{t_{i+1}})$ is estimated as:
    \begin{equation}
        \begin{aligned}
        &\nabla_\theta \mathcal{D}\bigl(\mathrm{sg}\bigl(\vf_\theta(\rvx_{t_{i}}, \sigma_{t_i})\big),\vf_\theta(\rvx_{t_{i+1}},\sigma_{t_{i+1}})\bigr) \\
        &=  (1 + o(1)) C\alpha
        \left[ \| \partial_\mathrm{CT}\vf_\theta \|^{\alpha - 2} (\partial_\mathrm{CT}\vf_\theta)^\top \frac{\partial \vf_\theta}{\partial \theta} \right] \cdot (t_{i+1} - t_i)^{\alpha-1}
        \end{aligned}
    \end{equation}
    Combining these two estimates, we can now compute the limit of the scaled gradient $N^{\alpha-1} \nabla_\theta \Ls_\bullet(\theta)$ for each $\bullet \in \{ \mathrm{CD}, \mathrm{CT} \}$ as:
    \begin{align}
        &N^{\alpha-1} \nabla_\theta \Ls_\bullet(\theta) \nonumber \\
        &= C\alpha T^{\alpha-2} \sum_{i=0}^{N-1} \lambda(\sigma_{t_i}) \E \biggl[  (1 + o(1)) \left[ \left\| \partial_\bullet \vf_\theta \right\|^{\alpha-2} ( \partial_\bullet \vf_\theta )^\top \frac{\partial \vf_\theta}{\partial \theta} \right] \biggr] \cdot (t_{i+1} - t_i) \\
        &\to C\alpha T^{\alpha-2} \int_{0}^{T} \lambda(\sigma_t) \E \biggl[ \left\| \partial_\bullet \vf_\theta (\rvx_t, \sigma_t) \right\|^{\alpha-2} \left( \partial_\bullet \vf_\theta (\rvx_t, \sigma_t) \right)^\top \frac{\partial \vf_\theta}{\partial \theta} (\rvx_t, \sigma_t) \biggr] \, \dif t
    \end{align}
    as $N \to \infty$. Finally, if $\alpha \neq 2$, then the term $\left\| \partial_\bullet \vf_\theta \right\|^{\alpha-2} \partial_\bullet \vf_\theta^\top$ is a nonlinear transformation of $\partial_\bullet \vf_\theta$. This nonlinearity tells that, in general,
    \begin{align}
        \E\left[ \left\| \partial_\text{CT} \vf_\theta \right\|^{\alpha-2} \left(\partial_\text{CT} \vf_\theta\right)^\top \middle| \rvx_t \right]
        \neq \left\| \partial_\text{CD} \vf_\theta \right\|^{\alpha-2} \left(\partial_\text{CD} \vf_\theta\right)^\top.
        \label{eq:limit_gradient_gap}
    \end{align}
    Therefore, the scaled gradient limits are not identical as functions of $\theta$, and in particular, their zero sets do not coincide.
\end{proof}

\paragraph{Differences with \citet{song23consistency}'s results. }
The previous theorem states a discrepancy between CT and CD objectives. However, \citet{song23consistency} provide equivalence results between consistency training and consistency distillation.  The differences come from the following reasons.

\begin{itemize}
    \item In  \citet[Theorem 2]{song23consistency}, it is stated that $L_{\mathrm{CT}} = L_{\mathrm{CD}}+o(\Delta T)$. However, in this theorem, the $o(\Delta T)$ is actually too large compared to the other term, and consequently the result is uninformative. Indeed it has two the two following problems: (i) if the distance function decays faster than the norm does, i.e., $\mathcal{D}(x, y) = o(\|x - y\|)$, then the $o(\Delta T)$ term is actually too large compared to the magnitude of the two losses as $N \to \infty$; (ii) The $C^2$-regularity assumption on the distance function $\mathcal{D}$ is too restrictive, excluding many cases such as the distance function given by a norm. For example, such a breakdown happens when $\mathcal{D}(x, y)$ is a metric induced by the norm, i.e., $\mathcal{D}(x, y) = \|x - y\|$. In this case, its partial derivatives, such as $\partial_2 \mathcal{D}(x, y) = \frac{\partial}{\partial y} \mathcal{D}(x, y)$ appearing in the proof, are undefined along x = y.
    \item  In \citet[Theorem 6]{song23consistency}, the theorem about limiting gradient equality is stated with a general distance function $\mathcal{D}$. However, the requirements on the Hessian of the distance function restrict the theorem's validity where the distance function is an (asymptotic) quadratic loss. Indeed, in their proof, it turns out that the Hessian can define a non-zero value only when $\mathcal{D}$ is an (asymptotic) quadratic loss. This coincides with our results in the case $\alpha=2$. 
\end{itemize}

\subsection{Proxy of the Regularizer}
\label{app:proof_proxy}

In this subsection, we establish a theoretical result about the decay rate of the proxy of the regularizer. As preparation for the main result and for future use, we introduce a simple lemma that decomposes the forward flow generated by a vector field into the sum of a scaling term and a correction term that is well-behaved.

\begin{lemma}\label{lemma:forward_flow}
    Assume that $\boldsymbol{\phi}$ is the forward flow generated by the vector field $\rvv_t$, meaning that it solves the characteristic equation:
    \begin{equation}
        \frac{\partial}{\partial t} \boldsymbol{\phi}(\rvx, \sigma_t) 
        = \rvv_t(\boldsymbol{\phi}(\rvx, \sigma_t)), \qquad
        \boldsymbol{\phi}(\rvx, \sigma_0) = \rvx.
    \end{equation}
    Also, assume that $\rvv_t$ is defined as
    \begin{equation}\label{eq:vector_field_from_denoiser}
        \rvv_t(\rvx) = \frac{\dot{\sigma}_t}{\sigma_t} (\rvx - \boldsymbol{D}(\rvx, \sigma_t))
    \end{equation}
    for some function $\boldsymbol{D}$, which we call a ``denoiser''. Then $\boldsymbol{\phi}$ satisfies the following integral equation:
    \begin{equation}
        \rvx
        = \frac{\sigma_0}{\sigma_t}\boldsymbol{\phi}(\rvx, \sigma_t) + \sigma_0 \int_{0}^{t} \frac{\dot{\sigma}_s}{\sigma_s^2} \boldsymbol{D}(\boldsymbol{\phi}(\rvx, \sigma_s), \sigma_s) \, \dif s.
    \end{equation}
\end{lemma}

\begin{proof}
    We first compute the derivative of $\boldsymbol{\phi}/\sigma_t$:
    \begin{align}
        \frac{\partial}{\partial t} \left( \frac{\boldsymbol{\phi}(\rvx, \sigma_t)}{\sigma_t} \right)
        &= -\frac{\dot{\sigma}_t}{\sigma_t^2} \boldsymbol{\phi}(\rvx, \sigma_t)
        + \frac{1}{\sigma_t} \cdot \frac{\dot{\sigma}_t}{\sigma_t} (\boldsymbol{\phi}(\rvx, \sigma_t) - \boldsymbol{D}(\boldsymbol{\phi}(\rvx, \sigma_t), \sigma_t)) \\
        &= -\frac{\dot{\sigma_t}}{\sigma_t^2} \boldsymbol{D}(\boldsymbol{\phi}(\rvx, \sigma_t), \sigma_t). \label{eq:scaled_characteristic}
    \end{align}
    Integrating both sides with respect to $t$, it follows that
    \begin{equation}
        \frac{\boldsymbol{\phi}(\rvx, \sigma_t)}{\sigma_t} - \frac{\boldsymbol{\phi}(\rvx, \sigma_0)}{\sigma_0}
        = - \int_{0}^{t} \frac{\dot{\sigma_s}}{\sigma_s^2} \boldsymbol{D}(\boldsymbol{\phi}(\rvx, \sigma_s), \sigma_s) \, \dif s.
    \end{equation}
    Rearranging and applying the initial condition $\boldsymbol{\phi}(\rvx, \sigma_0) = \rvx$, we obtain the desired equation.
\end{proof}

As an immediate consequence of this lemma, we obtain the following result about the asymptotic structure of a trained consistency model:

\begin{lemma}\label{lemma:cm_almost_scaling}
    Assume that $\mathring{\vf}$ is the consistency model generated by a bounded denoiser $\boldsymbol{D}$, in the sense that $\mathring{\vf}$ solves the transport equation
    \begin{equation}
        \frac{\partial \mathring{\vf}}{\partial \sigma}(\rvx, \sigma_t) \dot{\sigma}_t + \frac{\partial \mathring{\vf}}{\partial \rx}(\rvx, \sigma_t) \cdot \rvv_t(\rvx) = 0
    \end{equation}
    for a vector field $\mathring{\rvv}_t$ defined as in \cref{eq:vector_field_from_denoiser} with the denoiser $\boldsymbol{D}$. Then
    \begin{equation}
        \mathring{\vf}(\rvx, \sigma_t) = \frac{\sigma_0}{\sigma_t} \rvx + \mathcal{O}(1)
    \end{equation}
    uniformly in $\rvx$ and $\sigma_t$. The implicit bound of the error term can be chosen to be the bound of $\boldsymbol{D}$.
\end{lemma}

\begin{proof}
    Let $\boldsymbol{\phi}$ be the forward flow generated by $\mathring{\rvv}_t$ as in Lemma \ref{lemma:forward_flow}. This $\boldsymbol{\phi}$ is precisely the inverse of the consistency model $\mathring{\vf}$, in the sense that $\boldsymbol{\phi}(\mathring{\vf}(\rvx, \sigma), \sigma) = \rvx$ holds. Then, replacing $\rvx$ in the equation of Lemma \ref{lemma:forward_flow} with $\mathring{\vf}(\rvx, \sigma_t)$, we get
    \begin{equation}
        \mathring{\vf}(\rvx, \sigma_t)
        = \frac{\sigma_0}{\sigma_t}\rvx + \sigma_0 \int_{0}^{t} \frac{\dot{\sigma}_s}{\sigma_s^2} \boldsymbol{D}(\boldsymbol{\phi}(\mathring{\vf}(\rvx, \sigma_t), \sigma_s), \sigma_s) \, \dif s.
        \label{eq:consistency_model_error}
    \end{equation}
    Now let $R$ be such that $\|\boldsymbol{D}(\rvx, \sigma)\| \leq R$ for any $\rvx \in \mathds{R}^d$ and noise level $\sigma$. Then, the integral term in \cref{eq:consistency_model_error} is bounded as:
    \begin{equation}
        \left\| \sigma_0 \int_{0}^{t} \frac{\dot{\sigma}_s}{\sigma_s^2} \boldsymbol{D}(\boldsymbol{\phi}(\mathring{\vf}(\rvx, \sigma_t), \sigma_s), \sigma_s) \, \dif s \right\|
        \leq \sigma_0 \int_{0}^{t} \frac{\dot{\sigma}_s}{\sigma_s^2} R \, \dif s
        = \sigma_0 R \left( \frac{1}{\sigma_0} - \frac{1}{\sigma_t} \right)
        \leq R.
    \end{equation}
    This proves the desired claim.
\end{proof}

Now we turn to the main result, which analyzes the asymptotic behavior of $\tilde{\mathcal{R}}_{t,\mathrm{IC}}$ and $\tilde{\mathcal{R}}_{t,\mathrm{GC}}$, as $t \to \infty$:

\begin{reptheorem}{theorem_R_IC_GC}
    Assume that the data distribution contains more than a single point. Also, assume that the generator-augmented coupling between the predicted data point $\hat{\rvx}_t$ and noise $\rvz$ is computed via an ideal consistency model $\mathring{\vf}$, \textit{i.e.}, the flow of the PF-ODE. Then, as $t \to \infty$,
    \begin{equation}
        \tilde{\mathcal{R}}_{t,\mathrm{GC}} \ll \tilde{\mathcal{R}}_{t,\mathrm{IC}}. 
    \end{equation}
\end{reptheorem}

\begin{proof}
    We first investigate the asymptotic behavior of $\tilde{\mathcal{R}}_{t,\mathrm{IC}}$ in the limit of $t \to \infty$. Recall that the diffusion process $\rvx_t$ is given by $\rvx_t = \rvx_\star + \sigma_t \rvz$ for $(\rvx_\star, \rvz) \sim q_\text{I}$, and note that
    \begin{equation}
        \dot{\rvx}_t - \rvv_t(\rvx_t)
        = \dot{\sigma}_t \rvz - \E[ \dot{\sigma}_t \rvz | \rvx_t]
        = - \frac{\dot{\sigma}_t}{\sigma_t} ( \rvx_\star - \boldsymbol{D}(\rvx_t, \sigma_t) ),
    \end{equation}
    where $\boldsymbol{D}(\rvx_t, \sigma_t) = \E[\rvx_\star | \rvx_t]$ is the denoiser. Plugging this into the definition of $\tilde{\mathcal{R}}_{t,\mathrm{IC}}$, we get
    \begin{equation}
        \tilde{\mathcal{R}}_{t,\mathrm{IC}}
        = \left( \frac{\dot{\sigma}_t}{\sigma_t} \right)^2 \E\left[ \left\|  \rvx_\star - \boldsymbol{D}(\rvx_t, \sigma_t) \right\|^2 \right].
    \end{equation}
    Now, we claim that $\boldsymbol{D}(\rvx_t, \sigma_t) = \E[\rvx_\star | \rvx_t] \to \E[\rvx_\star]$ as $t \to \infty$. Intuitively, this is because $\rvx_t \approx \sigma_t \rvz$ for large $t$, and $\sigma_t \rvz$ is independent of $\rvx_\star$. More formally, note that the conditional distribution of $\rvx_t$ given $\rvx_\star$ is $p(\rvx_t | \rvx_\star) = \mathcal{N}(\rvx_t ; \rvx_\star, \sigma_t^2 \mathbf{I})$. By Bayes' theorem, the conditional distribution of $\rvx_\star$ given $\rvx_t$ is
    \begin{equation}
       p(\rvx_\star | \rvx_t)
        = \frac{p(\rvx_t | \rvx_\star)p(\rvx_\star)}{\int_{\mathds{R}^d} p(\rvx_t | \rvx'_\star)p(\rvx'_\star) \, \dif \rvx'_\star}
        = \frac{\exp\left(-\frac{1}{2\sigma_t^2}\left| \rvx_t - \rvx_\star \right|^2 \right) p(\rvx_\star)}{\int_{\mathds{R}^d} \exp\left(-\frac{1}{2\sigma_t^2}\left| \rvx_t - \rvx'_\star \right|^2 \right) p(\rvx'_\star) \, \dif \rvx'_\star}.
    \end{equation}
    As $t \to \infty$, we have $\sigma_t \to \infty$, so the exponential terms converge to $1$. Consequently, $p(\rvx_\star | \rvx_t) \to p(\rvx_\star)$ and hence $\E[\rvx_\star | \rvx_t] \to \E[\rvx_\star]$ as claimed. Thus,
    \begin{equation}
        \tilde{\mathcal{R}}_{t,\mathrm{IC}}
        \sim \left( \frac{\dot{\sigma}_t}{\sigma_t} \right)^2 \E\left[ \left\|  \rvx_\star - \E[\rvx_\star] \right\|^2 \right].
    \end{equation}
    Since the data distribution $p_\star$ is assumed to have more than one point, the variance $\E[ \|  \rvx_\star - \E[\rvx_\star] \|^2 ]$ is strictly positive. Therefore, $\tilde{\mathcal{R}}_{t,\mathrm{IC}}$ decays at a rate asymptotically proportional to $(\frac{\dot{\sigma}_t}{\sigma_t})^2$.

    Next, we investigate the asymptotic behavior of $\tilde{\mathcal{R}}_{t,\text{GC}}$. Recall the consistency training loss for GC, \cref{eq:consistency_gc_loss}. Under the assumptions in Theorem \ref{theorem_R_theta}, the scaled loss $N^{\alpha}\Ls_\text{GC}(\theta) $ converges to
    \begin{equation}\label{eq:limit_loss_gc}
        \Ls^{\infty}_\mathrm{GC}(\theta)
        = CT^{\alpha-1} \int_{0}^{T} \lambda(\sigma_t) \E \left[ \left\| \frac{\partial \vf_\theta}{\partial \sigma}(\tilde{\rvx}_t, \sigma_t) \dot{\sigma}_t + \frac{\partial \vf_\theta}{\partial \rvx}(\tilde{\rvx}_t, \sigma_t) \cdot \dot{\sigma}_t \rvz \right\|^{\alpha} \right] \, \dif t.
    \end{equation}
    Here, $\tilde{\rvx}_t = \hat{\rvx}_t + \sigma_t \rvz$ and $\hat{\rvx}_t = \mathring{\vf}(\rvx_t, \sigma_t)$, where $\mathring{\vf}$ is the ideal consistency model for the flow associated with the diffusion process $\rvx_t$. The proof of this claim is similar to that of Theorem~\ref{theorem_R_theta}, so we only highlight the necessary changes. Most importantly, the velocity term is not $\dot{\tilde{\rvx}}_t$ but $\dot{\sigma}_t \rvz$. This is due to how the discrete-time samples are constructed. Indeed, from \cref{eq:intermediate_points_GI}, we find that $\tilde{\rvx}_{t_{i+1}} - \tilde{\rvx}_{t_i} = (\sigma_{t_{i+1}} - \sigma_{t_i}) \rvz$, which manifests as the velocity term $\dot{\sigma}_t \rvz$ in \cref{eq:limit_loss_gc}. Consequently, the associated (average) velocity field $\tilde{\rvv}_t$ is given by
    \begin{equation}
        \tilde{\rvv}_t(\tilde{\rvx}_t)
        = \E[ \dot{\sigma}_t \rvz | \tilde{\rvx}_t]
        = \frac{\dot{\sigma}_t}{\sigma_t} (\tilde{\rvx}_t - \E[ \hat{\rvx}_t | \tilde{\rvx}_t]).
    \end{equation}
    Therefore, $\tilde{\mathcal{R}}_{t,\text{GC}}$ reduces to
    \begin{equation}
        \tilde{\mathcal{R}}_{t,\mathrm{GC}}
        = \left( \frac{\dot{\sigma}_t}{\sigma_t} \right)^2 \E\left[ \left\|  \hat{\rvx}_t - \E[ \hat{\rvx}_t | \tilde{\rvx}_t] \right\|^2 \right].
    \end{equation}
    Now, unlike in the IC case, we claim that $\E[ \hat{\rvx}_t | \tilde{\rvx}_t] \approx \hat{\rvx}_t$ as $t \to \infty$. Heuristically, this is because both $\hat{\rvx}_t$ and $\tilde{\rvx}_t$ are almost deterministic functions of $\rvz$; hence, the conditioning has negligible effect in the limit.
    
    More precisely, let $\boldsymbol{\phi}$ be the forward flow generated by the PF-ODE vector field $\rvv_t$. As in the proof of Lemma \ref{lemma:forward_flow}, integrating both sides of \cref{eq:scaled_characteristic} from $t$ to $u$ yields
    \begin{equation}
        \frac{\boldsymbol{\phi}(\rvx, \sigma_u)}{\sigma_u}
        = \frac{\boldsymbol{\phi}(\rvx, \sigma_t)}{\sigma_t} - \int_{t}^{u} \frac{\dot{\sigma_s}}{\sigma_s^2} \boldsymbol{D}(\boldsymbol{\phi}(\rvx, \sigma_s), \sigma_s) \, \dif s.
    \end{equation}
    Letting $u \to \infty$, we claim that the right-hand side converges. Indeed, the empirical data distribution $p_\star$ has compact support, meaning all the data points are confined in a bounded region of $\mathds{R}^d$. Since the values of $\boldsymbol{D}$ are weighted averages of the data points, it follows that $\boldsymbol{D}$ is also bounded. Then the integrand $\frac{\dot{\sigma_s}}{\sigma_s^2} \boldsymbol{D}(\boldsymbol{\phi}(\rvx, \sigma_s), \sigma_s)$ is absolutely integrable on $[t, \infty)$, hence the convergence follows. Moreover, the limit does not depend on $t$. Denote this limit by $\rho(\rvx)$:
    \begin{equation}\label{eq:forward_flow_limit}
        \boldsymbol{\rho}(\rvx) = \frac{\boldsymbol{\phi}(\rvx, \sigma_t)}{\sigma_t} - \int_{t}^{\infty} \frac{\dot{\sigma_s}}{\sigma_s^2} \boldsymbol{D}(\boldsymbol{\phi}(\rvx, \sigma_s), \sigma_s) \, \dif s.
    \end{equation}
    As shown in the previous part, we know that $\boldsymbol{D}(\rvx, t) = c + o(1)$ as $t\to\infty$ with $c = \E[x_\star]$. Then, multiplying both sides of \cref{eq:forward_flow_limit} by $\sigma_t$ and rearranging, we have, for large $t$,
    \begin{align}
        \boldsymbol{\phi}(\rvx, \sigma_t)
        &= \sigma_t \boldsymbol{\rho}(\rvx) + \sigma_t \int_{t}^{\infty} \frac{\dot{\sigma_s}}{\sigma_s^2} \boldsymbol{D}(\boldsymbol{\phi}(\rvx, \sigma_s), \sigma_s) \, \dif s \\
        &= \sigma_t \boldsymbol{\rho}(\rvx) + (c + o(1)) \sigma_t \int_{t}^{\infty} \frac{\dot{\sigma_s}}{\sigma_s^2} \, \dif s \\
        &= \sigma_t \boldsymbol{\rho}(\rvx) + c + o(1).
    \end{align}
    Since $\boldsymbol{\phi}$ is a bijection, the above relation tells that $\boldsymbol{\rho}(\rvx)$ is also a bijection. Next, we replace $\rvx \leftarrow \hat{\rvx}_t$ in the equation defining $\boldsymbol{\rho}(\rvx)$, \cref{eq:forward_flow_limit}, to obtain:
    \begin{equation}
        \boldsymbol{\rho}(\hat{\rvx}_t) = \rvz + \frac{\rvx_\star}{\sigma_t} - \int_{t}^{\infty} \frac{\dot{\sigma_s}}{\sigma_s^2} \boldsymbol{D}(\boldsymbol{\phi}(\hat{\rvx}_t, \sigma_s), \sigma_s) \, \dif s.
    \end{equation}
    Since $\boldsymbol{\rho}$ is invertible, applying $\boldsymbol{\rho}^{-1}$ to both sides yields
    \begin{align}
        \hat{\rvx}_t
        &= \boldsymbol{\rho}^{-1}\left( \rvz + \frac{\rvx_\star}{\sigma_t} - \int_{t}^{\infty} \frac{\dot{\sigma_s}}{\sigma_s^2} \boldsymbol{D}(\boldsymbol{\phi}(\hat{\rvx}_t, \sigma_s), \sigma_s) \, \dif s \right) \\
        &= \boldsymbol{\rho}^{-1}\left( \frac{\tilde{\rvx}_t}{\sigma_t} + \frac{\rvx_\star - \hat{\rvx}_t}{\sigma_t} - \int_{t}^{\infty} \frac{\dot{\sigma_s}}{\sigma_s^2} \boldsymbol{D}(\boldsymbol{\phi}(\hat{\rvx}_t, \sigma_s), \sigma_s) \, \dif s \right)
    \end{align}
    Since all of $\rvx_\star$, $\hat{\rvx}_t$, and $\boldsymbol{D}$ are bounded by the largest norm of the data point, they are all finite. Hence, the last line shows that $\hat{\rvx}_t = \boldsymbol{\rho}^{-1} \bigl( \frac{\tilde{\rvx}_t}{\sigma_t} + \mathcal{O}(\frac{1}{\sigma_t}) \bigr)$, demonstrating that $\hat{\rvx}_t$ is almost a deterministic function of $\tilde{\rvx}_t$. Therefore, $\E[\hat{\rvx}_t | \tilde{\rvx}_t] \approx \hat{\rvx}_t$ as required. Consequently, $\tilde{\mathcal{R}}_{t,\mathrm{GC}}$ satisfies
    \begin{equation}
        \tilde{\mathcal{R}}_{t,\mathrm{GC}}
        \ll \left( \frac{\dot{\sigma}_t}{\sigma_t} \right)^2.
    \end{equation}
    This proves that $\tilde{\mathcal{R}}_{t,\mathrm{GC}} \ll \tilde{\mathcal{R}}_{t,\mathrm{IC}}$ as required.
\end{proof}

\subsection{Transport Cost}
\label{app:proof_transport}

As a base for the two corollaries presented in the paper, we will first derive a useful representation of the derivative of the transport cost.

The main purpose of the lemma is to provide a more tractable representation of $c'(t)$, the time derivative of the expected transport cost. We expect $c(t)$ to decrease with $t$ because the predicted data point $\mathring{\vf}(\mathbf{x}_t, \sigma_t)$ becomes more dependent on the noise $\mathbf{z}$ as $t$ increases. However, directly analyzing $\mathring{\vf}(\mathbf{x}_t, \sigma_t) - \mathbf{z}$ is challenging because the dependence of $\mathring{\vf}(\mathbf{x}_t, \sigma_t)$ on $\mathbf{z}$ is not explicit. Therefore, the lemma aims to:

\begin{itemize}
    \item identify a quantity that better captures the dependence between $\mathbf{z}$ and $\mathbf{x}_t$;
    \item relate $c(t)$ to this quantity.
\end{itemize}

The proof proceeds by deriving a key property of the ground-truth consistency map $\mathring{\vf}$: it satisfies the transport equation,

\begin{equation}
    \frac{\partial \mathring{\vf}}{\partial \sigma}(\mathbf{x}, \sigma_t) \, \dot{\sigma}_t + \frac{\partial \mathring{\vf}}{\partial \mathbf{x}}(\mathbf{x}, \sigma_t) \cdot \mathbf{v}_t(\mathbf{x}) = 0.
\end{equation}

This equation is equivalent to saying that the conditional expectation of the time derivative of $\mathring{\vf}(\mathbf{x}_t, \sigma_t)$ given $\mathbf{x}_t$ is zero:

\begin{equation}
\mathbb{E}\left[ \frac{\partial}{\partial t} \mathring{\vf}(\mathbf{x}_t, \sigma_t) \,\middle|\, \mathbf{x}_t \right] = 0.
\end{equation}

By leveraging this property, we can simplify $c'(t)$ into an expression involving $\mathbf{w}_t = \mathbf{z} - \mathbb{E}[\mathbf{z} \mid \mathbf{x}_t]$, the residual between the true noise $\mathbf{z}$ and its prediction given $\mathbf{x}_t$. This residual captures the uncertainty in predicting $\mathbf{z}$ based on $\mathbf{x}_t$, allowing us to relate $c'(t)$ directly to the prediction accuracy of $\mathring{\vf}$.

\begin{replemma}{lemma:transport_cost}[\textbf{Transport cost of GC coupling}]
    Assume that $\mathring{\vf}$ is a continuously differentiable function representing the ground-truth consistency model, \textit{i.e.}\ the flow of the PF-ODE induced by the diffusion process $\rvx_t$. Define $\rvw_t = \rvz - \E[\rvz | \rvx_t] = \frac{1}{\dot{\sigma}_t} (\dot{\rvx}_t - \E[\dot{\rvx}_t \mid \rvx_t])$. Then:
    \begin{align}
        c'(t) &= - 2 \dot{\sigma}_t \E\left[ \left< \frac{\partial \mathring{\vf}}{\partial \rvx} (\rvx_t, \sigma_t) \cdot \rvw_t, \rvw_t \right> \right].
    \end{align}
\end{replemma}

\begin{proof}
Note that the inverse flow $\mathring{\vf}^{-1}(\rvy, \sigma_t)$ transports the initial point $\rvy$ at time $t = 0$ along the vector field $\rvv_t$ up to time $t$. Consequently, $\mathring{\vf}^{-1}$ is a flow with the corresponding vector field $\rvv_t$:
\begin{equation}
    \frac{\partial }{\partial t} [\mathring{\vf}^{-1}(\rvy, \sigma_t)] = \rvv_t(\mathring{\vf}^{-1}(\rvy, \sigma_t)).
\end{equation}
By differentiating both sides of the identity $\rvy = \mathring{\vf}(\mathring{\vf}^{-1}(\rvy, \sigma_t), \sigma_t)$ with respect to $t$ and applying the above observation, we get:
\begin{align}
    0 &= \frac{\partial}{\partial t} \left[ \mathring{\vf}(\mathring{\vf}^{-1}(\rvy, \sigma_t), \sigma_t) \right] \\
    &= \frac{\partial \mathring{\vf}}{\partial \sigma} (\mathring{\vf}^{-1}(\rvy, \sigma_t), \sigma_t) \dot{\sigma}_t + \frac{\partial \mathring{\vf}}{\partial \rvx} (\mathring{\vf}^{-1}(\rvy, \sigma_t), \sigma_t) \cdot \frac{\partial}{\partial t} [\mathring{\vf}^{-1}(\rvy, \sigma_t)]\\
    &= \frac{\partial \mathring{\vf}}{\partial \sigma}(\rvx, \sigma_t) \dot{\sigma}_t + \frac{\partial \mathring{\vf}}{\partial \rvx}(\rvx, \sigma_t) \cdot \rvv_t(\rvx),
\end{align}
where the substitution $\rvx = \mathring{\vf}^{-1}(\rvy, \sigma_t)$ is used in the last step. Consequently,
\begin{align}
    c'(t)
    &= 2 \E\left[ \left\langle 
        \frac{\partial}{\partial t}[\mathring{\vf}(\rvx_t, \sigma_t)], 
        \mathring{\vf}(\rvx_t, \sigma_t) - \rvz
    \right\rangle \right] \\
    &= 2 \E\left[ \left\langle 
        \frac{\partial \mathring{\vf}}{\partial \sigma}(\rvx_t, \sigma_t) \dot{\sigma}_t + \frac{\partial \mathring{\vf}}{\partial \rvx}(\rvx_t, \sigma_t) \cdot \dot{\rvx}_t, 
        \mathring{\vf}(\rvx_t, \sigma_t) - \rvz
    \right\rangle \right] \\
    &= 2 \E\left[ \left\langle 
        \frac{\partial \mathring{\vf}}{\partial \rvx}(\rvx_t, \sigma_t) \cdot (\dot{\rvx}_t - \rvv_t(\rvx)), 
        \mathring{\vf}(\rvx_t, \sigma_t) - \rvz
    \right\rangle \right] \\
    &= 2\dot{\sigma}_t \E\left[ \left\langle 
        \frac{\partial \mathring{\vf}}{\partial \rvx}(\rvx_t, \sigma_t) \cdot (\rvz - \E[\rvz | \rvx_t]), 
        \mathring{\vf}(\rvx_t, \sigma_t) - \rvz
    \right\rangle \right],
\end{align}
where we used the relations $\rvx_t = \rvx_\star + \sigma_t \rvz$ and $\rvv_t(\rvx) = \E[\dot{\rvx}_t | \rvx_t]$. Now, let $\rvw_t = \rvz - \E[\rvz \mid \rvx_t]$. Then $\E[\rvw_t \mid \rvx_t] = 0$, hence by an application of the law of iterated expectations, $\E[ \langle \mathbf{w}_t, g(\rvx_t) \rangle ] = 0$ for essentially any function $g : \mathbb{R}^d \to \mathbb{R}^d$. Using this, we can further simplify the last line as:
\begin{align}
    c'(t)
    &= - 2 \dot{\sigma}_t \E\left[ \left\langle
         \frac{\partial \mathring{\vf}}{\partial \rvx}(\rvx_t, \sigma_t) \cdot \mathbf{w}_t, 
         \rvz 
     \right\rangle \right]
     = - 2 \dot{\sigma}_t \E\left[ \left\langle
         \frac{\partial \mathring{\vf}}{\partial \rvx}(\rvx_t, \sigma_t) \cdot \mathbf{w}_t, 
         \mathbf{w}_t
     \right\rangle \right],
\end{align}
proving the desired equality.
\end{proof}

An immediate consequence of this lemma is that $c(t)$ decreases for small $t$:

\begin{repcorollary}{cor:transport_t0}[\textbf{Decreasing transport cost of GC coupling in $t\to 0^+$}]
    There exists a $t_* > 0$ such that for all $t \in [0,t_*]$, the derivative of $c(t)$ takes the form $  c'(t) = - 2 \dot{\sigma}_t a_t$ with $a_t > 0$. Hence for $\dot{\sigma}_t$  positive, the cost is decreasing.  In particular, in the EDM setting where $\sigma_t = t$, $c(t)$ is decreasing for small $t$.
\end{repcorollary}

\begin{proof}    
    The proof of this corollary proceeds by noting that for $t=0$, the consistency model $\mathring{\vf}(\rvx,t)$ is an identity function, its Jacobian is an identity matrix leading to  $a_t = \E [ \|\rvw_t\| ^2] > 0$  and by assumption, all the elements of the Jacobian are continuous. By continuity of $a_t$, $t_*$ exists and invoking intermediate value theorem on $a_t$ concludes the proof.
\end{proof}

The next result is the statement about the asymptotic behavior of the transport cost $c(t)$ in the large-$t$ regime.

\begin{repcorollary}{cor:transport_tmax}[\textbf{Decreasing transport cost of GC coupling in $t \approx t_{\max}$}]
    Assume that the consistency model $\mathring{\vf}(x, \sigma)$ is a scaling function $\mathring{\vf}(\rvx, \sigma_t) = \frac{\sigma_0}{\sigma_t} \rvx$. Then, we have $ c'(t) = - \frac{ 2 \dot{\sigma}_t \sigma_0}{ \sigma_t } \E[\| \rvw_t \|^2]$. In particular, $c(t)$ is decreasing whenever $\sigma_t$ is increasing.
\end{repcorollary}

\begin{proof}
    Under the assumption, we have $\frac{\partial \mathring{\vf}}{\partial \rvx}=\frac{\sigma_0}{\sigma_t}\mathbf{I}$. Thus, by Lemma \ref{lemma:transport_cost},
    \begin{equation}
        c'(t)
        = - 2 \dot{\sigma}_t \E\left[ \left\langle \frac{\sigma_0}{\sigma_t} \mathbf{I} \rvw_t, \rvw_t \right\rangle \right]
        = -\frac{2 \dot{\sigma}_t \sigma_0 }{\sigma_t} \E[\|\rvw_t\|^2].
    \end{equation}
    This proves that $c'(t) < 0$ whenever $\dot{\sigma}_t > 0$.
\end{proof}

\paragraph{Toy example.}
Let us consider a one-dimensional toy example where $\rvx_\star \sim \mathcal{N}(0,\sigma_\star^2)$ with $\sigma_\star \geq 0$ and $\rvz \sim \mathcal{N}(0, 1)$ are independent. Also, we assume $\sigma_0 = 0$ for the sake of simplicity. In this case, the marginal law of $\rvx_t$ is also Gaussian with $p_t = \mathcal{N}(0, \sigma_\star^2+\sigma_t^2)$, so the vector field for the diffusion process $\rvx_t$ is calculated as $\rvv_t(\rvx) = -\dot{\sigma}_t \sigma_t \nabla_\rvx \log p_t(\rvx) = \frac{\dot{\sigma}_t \sigma_t}{\sigma_\star^2+\sigma_t^2}\rvx $. Then, the corresponding target diffusion flow and the transport cost function are:
\begin{equation}
    \mathring{\vf}(\rvx, \sigma_t) = \frac{\sigma_\star}{\sqrt{ \sigma_\star^2 + \sigma_t^2 }} \, \rvx
    \quad\text{and}\quad
    c(t) = \sigma_\star^2 + 1 - \frac{2\sigma_\star \sigma_t}{\sqrt{\sigma_\star^2 + \sigma_t^2}}.
\end{equation}
We note that $\mathring{\vf}(\rvx, \sigma_t)$ is indeed a scaling function which is asymptotically proportional to $\frac{\rvx}{\sigma_t}$ for large $t$, and $c(t)$ is decreasing in $t$ for $t>0$.

\begin{figure*}
    \centering
    {
    \hfill
    \subfigure{\includegraphics[width=0.4\textwidth]{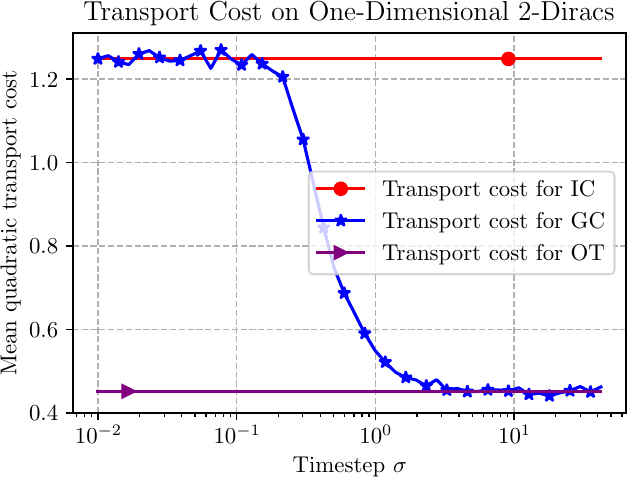}}
    \hfill
    \subfigure{\includegraphics[width=0.4\textwidth]{images/CIFAR10_EDM_transport_cost.pdf}}
    \hfill
    }
    \caption{
       Comparison of transport costs between IC, batch-OT, and GC on two 2D-Diracs (left) and CIFAR-10 (right). \label{fig:transport_cost_2d_Cifar} 
    }
\end{figure*}

\paragraph{Experimental validation. }
We validation the transport cost decrease in \cref{fig:transport_cost_2d_Cifar}, on a toy dataset composed of two 2D-Diracs, and on CIFAR-10. Interestingly, we observe that when computing OT transport plans between batches instead of on the full data, GC allows to reduce transport cost more than batch-OT. 

\subsection{Proxy Term} \label{subsec:proxy}

In this part, we clarify the connection between the proxy term and the  in the case of the quadratic loss ($\alpha=2$). Indeed, we can bound the regularization term with the proxy term thanks to the Jacobian's maximum singular value $s_{\text{max}}( \frac{\partial f_\theta}{\partial x} )$, which is bounded as typical networks are Lipschitz:

\begin{equation}
    \left\| \frac{\partial \boldsymbol{f}_\theta}{\partial \rvx}(\rvx_t, \sigma_t) \left( \dot{\rvx}_t - \rvv_t(\rvx_t) \right) \right\|^2 \leq \left\| \frac{\partial \boldsymbol{f}_\theta}{\partial x} \right\|^2 \|\dot{x}_t - \rvv_t(x_t)  \|^2 \leq s^2_{\text{max}} ( \frac{\partial \boldsymbol{f}_\theta}{\partial \rvx} )  \|\dot{\rvx}_t - \rvv_t(\rvx_t)  \|^2
\end{equation}

We could also use some assumptions on $\boldsymbol{f}$, \textit{e.g.} the fact that it is close to a scaling function for large $t$ (see Corrolary 2). If $\boldsymbol{f}(\rvx, \sigma_t) = \frac{\sigma_0}{\sigma_t} \rvx$, then we would have:

\begin{equation}
    \left\| \frac{\partial \boldsymbol{f}_\theta}{\partial \rvx}(\rvx_t, \sigma_t) \left( \dot{\rvx}_t - \rvv_t(\rvx_t) \right) \right\|^2 = (\frac{\sigma_0}{\sigma_t})^2 \|\dot{\rvx}_t - \rvv_t(\rvx_t)  \|^2.
\end{equation}

\section{Algorithm}
We present the detailed algorithm for GC ($\mu=\cdot$) in \cref{alg:consistency_model_w_GI}. 
\begin{algorithm}[tb]
   \caption{Training of consistency models with generator-augmented trajectories}
   \label{alg:consistency_model_w_GI}
\begin{algorithmic}
   \STATE {\bfseries Input:} Randomly initialized consistency model $\vf_\theta$, number of timesteps $N$, noise schedule $\sigma_{t_i}$, loss weighting $\lambda(\cdot)$, learning rate $\eta$, distance function $\mathcal{D}$, noise distribution $p_z$, joint learning parameter $\mu$.
   \STATE {\bfseries Output:} Trained consistency model $\vf_{\theta}$.
   \WHILE{not converged}
       \STATE $\rvx_\star \sim p_\star,~~\rvz \sim  p_z$ \hfill \COMMENT{batch of real data and noise vectors}
       \STATE $i \sim \text{multinomial}\big(p(\sigma_{t_0}),\ldots,p(\sigma_{t_N})\big)$ \hfill \COMMENT{sampling timesteps}
       \STATE $m \sim \text{binomial}(\mu, \text{size=batch\_size})$ \hfill \COMMENT{mask of dimension (batch\_size) with each $m_j \sim \text{binomial}(\mu)$}
       \STATE $\rvx_{t_{i}} \leftarrow \rvx_\star + \sigma_{t_i} \rvz$ \hfill \COMMENT{IC intermediate points}
       \STATE $\hat{\rvx}_{t_i} \leftarrow \mathrm{sg} \big(\vf_\theta(\rvx_{t_i}, \sigma_{t_i})\big)$ \hfill \COMMENT{endpoint prediction from the model}
       \STATE $\hat{\rvx}_{t_i} \leftarrow m \cdot \hat{\rvx}_{t_i} + (1-m) \cdot \rvx_\star$ \hfill \COMMENT{mixing IC and GC trajectories}
       \STATE $\tilde{\rvx}_{t_i} \leftarrow \hat{\rvx}_{t_i}  + \sigma_{t_i} \rvz$,~~$\tilde{\rvx}_{t_{i+1}} \leftarrow \hat{\rvx}_{t_i}  + \sigma_{t_{i+1}} \rvz$ \hfill \COMMENT{GC intermediate points}
       \STATE $\Ls(\theta) = \lambda( \sigma_{t_i} ) \mathcal{D}\big( \mathrm{sg}( \vf_\theta(\tilde{\rvx}_{t_i}, \sigma_{t_i}) ), \vf_\theta(\tilde{\rvx}_{t_{i+1}}, \sigma_{t_{i+1}}) \big)$ \hfill \COMMENT{consistency loss}
       \STATE $\theta \leftarrow \theta - \eta \nabla_\theta \Ls(\theta)$ \hfill \COMMENT{update model's weights}
   \ENDWHILE
\end{algorithmic}
\end{algorithm}

\section{Additional Results}

\subsection{Ablation Studies}
\begin{figure*}
     {\centering\hfill\subfigure[FID of IC vs GC models during training.]{
    \includegraphics[width=0.4\textwidth]{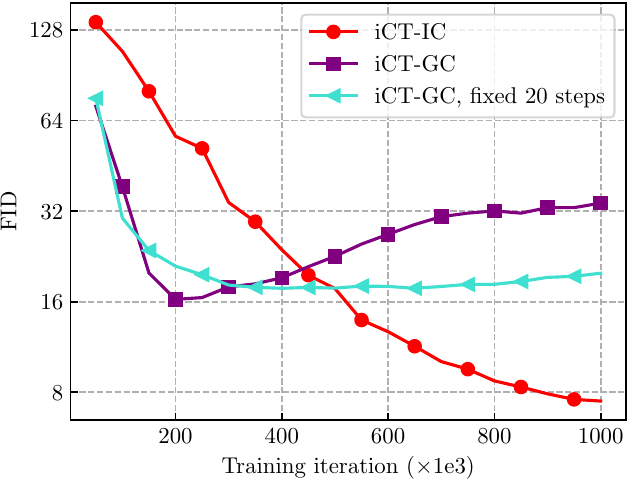}
    \label{fig:gi_scheduling}
}\hfill\subfigure[FID of trained IC vs GC along trajectories.]{
    \includegraphics[width=0.4\textwidth]{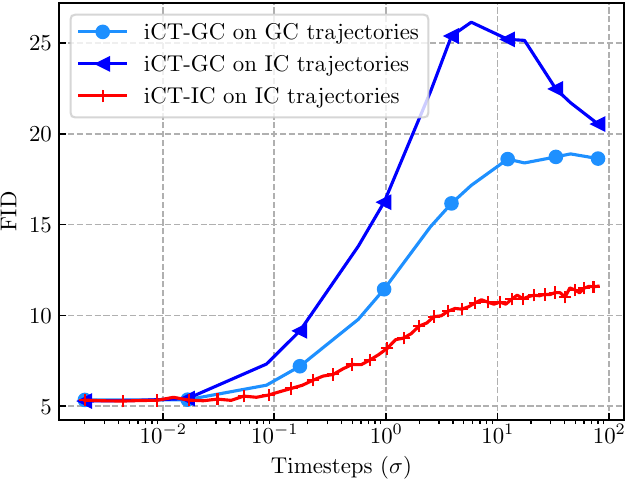}
    \label{fig:fid_per_timestep}
}
\hfill}
  \caption{Analysis of consistency models trained only with GC on CIFAR-10. (a) When trained with only GC trajectories, consistency models does not reach the performance of the base model (iCT-IC). In (b), we show that is linked to a distribution shift problem: GC models are weak on IC trajectoires, thus are sub-optimal for predicting $\hat{\rvx}_{t_i}$ required in their own training (\cref{eq:intermediate_points_IC}). \label{fig:GC_scratch}}
\end{figure*}

\label{sec:ablation}

\paragraph{Understanding why GC($\mu=1$) fails. \label{subsec:gc_alone}} 
This experiment involves training a consistency model with GC($\mu=1$). As shown in \cref{fig:gi_scheduling}, we observe that these models converge quickly but reach saturation early in the training process. When applying the timestep scheduling method with an increasing number of timesteps from \citet{song2023improved}, the FID of the models worsens. Using a fixed number of timesteps prevents divergence of the FID, but it still plateaus at a higher FID than iCT-IC.

In \cref{fig:fid_per_timestep}, we plot the FID per timestep for three model\,/\,trajectory pairs: GC($\mu=1$)-model on IC trajectories, GC($\mu=1$)-model on GC trajectories, and IC-model on IC trajectories. Notably, we observe a distribution shift between IC and GC trajectories: the FID of the GC-model on IC trajectories degrades at the intermediate timesteps of the diffusion process. This highlights why deviating from the theory and training a model exclusively on GC trajectories is insufficient:
to build $\rvx_{t_i}$ in \cref{eq:intermediate_points_IC}, the model is inferred on IC but trained on GC trajectories. If IC and GC differ too much, the model cannot improve on IC.


\begin{table}
    \caption{Analysis of performance with regards to some hyper-parameters of iCT-GC ($\mu=0.5$) on CIFAR-10.}\label{wrap-tab:ablation}
    \centering
        \begin{tabular}{lc}\\\toprule  
        Model & FID \\ \midrule
        iCT-IC & 7.42 $\pm$ 0.04 \\  
        iCT-GC $(\mu=0.5)$ iso-time & \textbf{6.38} $\pm$ 0.03 \\ 
        \midrule
        iCT-GC $(\mu=0.5)$ & \textbf{5.95} $\pm$ 0.05  \\  
        iCT-GC$(\mu=0.5)$ + dropout & {7.77} $\pm$ 0.04 \\ 
        iCT-GC $(\mu=0.5)$ - EMA  & {6.73} $\pm$ 0.05  \\ \bottomrule
        \end{tabular}
\end{table} 

\paragraph{Iso wall-clock training time. }
As illustrated above, consistency models trained with GC converge faster than IC. However, each training step is more time-consuming, as it necessitates a forward evaluation of the consistency model without gradient computation. Regarding wall-clock training time, the computational overhead of iCT-GC is approximately 20\% of the iCT-IC. In top part of \cref{wrap-tab:ablation}, we report under ``iCT-GC ($\mu=0.5$)  iso-time'' the results of iCT-GC ($\mu=0.5$) trained with the same wall-clock duration as iCT-IC. Even when considering wall-clock training time, iCT-GC ($\mu=0.5$) is still superior to iCT-IC. 

\paragraph{Hyper-parameters.}  We evaluate the influence of two important hyper-parameters. First, the dropout in the learned model. Second, whether to use or not the EMA to compute GC endpoints $\hat{\rvx}$. The results are presented in the bottom part of \cref{wrap-tab:ablation}. Interestingly, the results on dropout are opposite to those found by \citet{song2023improved}, since using dropout lowers the performance of iCT-GC ($\mu=0.5$).

\begin{figure}
    \centering
    {
    \subfigure{\includegraphics[width=0.5\textwidth]{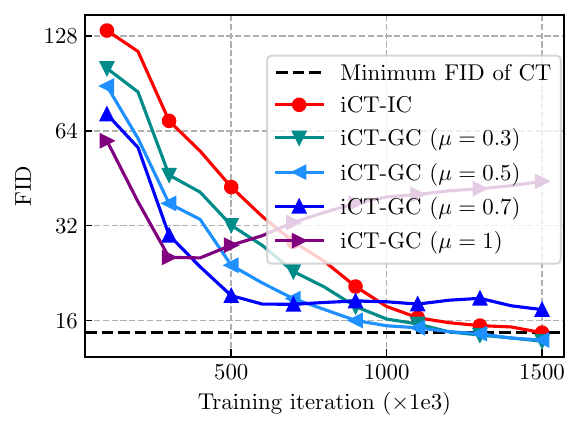}}}
    \caption{
       Results of varying $\mu$ for iCT-GC on ImageNet-32. \label{fig:mu_imagenet}
    }
\end{figure}

\paragraph{Analysis of $\mu$ on ImageNet.} We present further results of the joint learning procedure with varying $\mu$ ($\{0.3,0.5,0.7,1.\}$) on ImageNet-32 in \cref{fig:mu_imagenet}. For $\mu=\{0.3,0.5\}$, iCT-GC outperforms the base model iCT-IC.

\subsection{Visual Results}

\begin{figure}
    \centering
    {\hfill
    \subfigure[Trained with IC.]{\includegraphics[width=0.3\textwidth]{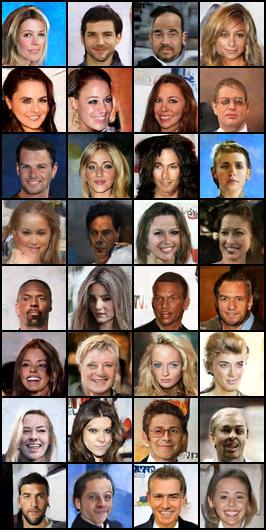}} 
    \hfill
    \subfigure[Trained with OT.]{\includegraphics[width=0.3\textwidth]{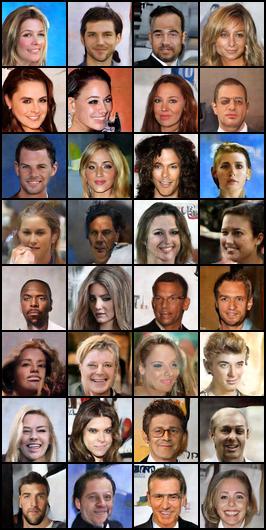}}
    \hfill
    \subfigure[Trained with GC ($\mu=0.5$).]{\includegraphics[width=0.3\textwidth]{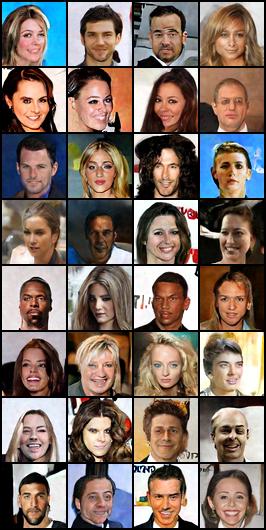}} 
    \hfill}
    \caption{
       Uncurated samples from consistency models trained on CelebA $64\times64$ for fixed noise vectors. Note that models trained with generator-augmented trajectories tend to generate sharper images. \label{fig:celeba_generated_img}
    }
\end{figure}

We include in \cref{fig:celeba_generated_img} examples of generated images for considered baselines.

\section{Experimental Details}
\label{app:details}

The code is based on the PyTorch library \citep{Paszke2019}. 

\paragraph{Scheduling functions and hyperparameters from \citet{song2023improved}.} The training of consistency models heavily rely on several scheduling functions. First, there is a noise schedule $\{\sigma_i\}_{i=0}^N$ which is chosen as in \citet{karras2022edm}. Precisely, 
$\sigma_i = \big( {\sigma_{\text{0}}}^{\frac{1}{\rho}} + \frac{i}{N} ( {\sigma_{\text{N}}}^{\frac{1}{\rho}} - {\sigma_{\text{0}}}^{\frac{1}{\rho}}  )   \big)^\rho$ 
with $\rho=7$. Second, there is a weighting function that affects the training loss, chosen as $\lambda(\sigma_i) = \frac{1}{\sigma_{i+1}-\sigma_i}$. Combined with the choice of noise schedule, it emphasizes to be consistent on timesteps with low noise. Then, \citet{song23consistency} propose to progressively increase the number of timesteps $N$ during training. \citet{song2023improved} argue that a good choice of dicretization schedule is an exponential one: $N(k) = \text{min}(s_0 2^{\lfloor \frac{k}{K'} \rfloor},s_1) + 1$ where $K'=\lfloor \frac{K}{\log_2[s_1/s_0] + 1}\rfloor$, $K$ is the total number of training steps, $k$ is the current  training step, $s_0$ (respectively $s_1$) the initial (respectively final) number of timesteps. Finally, \citet{song2023improved} propose a discrete probability distribution on the timesteps which mimics the continuous probability distribution recommended in the continuous training of score-based models by \citet{karras2022edm}. It is defined as $p(\sigma_i) \propto \text{erf}(\frac{\log(\sigma_{i+1}) - P_{\text{mean}}}{\sqrt{2} P_{\text{std}}}) - \text{erf}(\frac{\log(\sigma_{i}) - P_{\text{mean}}}{\sqrt{2} P_{\text{std}}})$. 
In practice, \citet{song2023improved} recommend using: $s_0=10$, $s_1=1280$, $\rho=7$, $P_{\text{mean}}=-1.1$, $P_{\text{std}}=2.0$.

We use the lion optimizer \citep{chen2024symbolic} implemented from \href{https://github.com/lucidrains/lion-pytorch}{https://github.com/lucidrains/lion-pytorch}.

\paragraph{Selection of hyper-parameter $\mu$.} We have selected $\mu$ based on the results from \cref{fig:mixing}, which presents a grid search for $\mu$ on CIFAR-10. Given the bell-shaped relationship observed between $\mu$ and FID, we opted to retain the best performing value identified on CIFAR-10, $\mu=0.5$, for all subsequent experiments (\cref{tab:full_results}), including those on other datasets, without further tuning. Importantly, even without an exhaustive hyperparameter search, our method consistently outperforms baseline approaches. This choice is validated by the ablation study presented in \cref{sec:ablation} showing similar trend for another dataset, showing that the hyper-parameter $\mu$ is easy to tune. 

In the ECT setting, we found that $\mu<0.5$ leads to improved performance, while $\mu>0.5$ can degrade final performance. Overall, we recommend setting $\mu$ to small values (around 0.3) since it leads to improved performance in all our experiments.  

\textbf{Details on neural networks architectures.} We use the NCSN++ architecture \citep{song2021scorebasedSDEs} and follow the implementation from \href{https://github.com/NVlabs/edm}{https://github.com/NVlabs/edm}. 

\textbf{Evaluation metrics.} We report the FID, KID and IS. For the three different metrics, we rely on the implementation from TorchMetrics \citep{SkafteDetlefsen2022}. For the three different metrics, we use the standard practice (e.g. \cite{song2023improved}) of FID which is to compare sets of \num{50000} generated versus training images. 
Confidence intervals reported in \cref{tab:full_results} are averaged on five runs by sampling new sets of training images, and new sets of generated images from the same model.

\textbf{Datasets. } CIFAR-10 is a dataset introduced in \citet{cifar}. ImageNet \citep{imagenet}, CelebA \citep{Liu2015}, and LSUN Church \citep{yu2015lsun} are used respectively at $32\times 32$, $64\times 64$ and $64\times 64$ resolutions. We preprocess these images by resizing smaller side to the desired value, center cropping, and linearly scaling pixel values to $[-1,1]$. 

\begin{table}
\centering
    \caption{Hyperparameters for CIFAR-10. Arrays indicate quantities per resolution of the UNet model. $\{\}$ indicate an hyper-parameter search performed for each type of model (iCT, iCT-OT, iCT-GC ($\mu=0.5$)).}
\label{tab:hyperparameters_cifar}
\vspace{\baselineskip}
\begin{tabular}{ll}
\toprule
\textbf{Hyperparameter} & \textbf{Value} \\
 \midrule
batch size & $512$ \\
image resolution & $32$ \\
training steps & \num{100000} \\
learning rate & $\{0.0001, 0.00003\}$ \\
optimizer & lion \\
$s_0$ & $10$ \\
$s_1$ & $1280$ \\
$\rho$ & $7$ \\
$\sigma_{0}$ & $0.002$ \\
$\sigma_{1}$  & $80$ \\
network architecture & SongUNet \\ & (from \cite{karras2022edm} implementation) \\
model channels & $128$ \\
dropout & $\{0.,0.3\}$ \\
num blocks & $3$ \\
embedding type & positional \\
channel multiplicative factor & $[1,2,2]$ \\
attn resolutions & $\emptyset$ \\
\bottomrule
\end{tabular}
\end{table}

\begin{table}
\centering
\caption{Hyperparameters for CelebA and LSUN Church. Arrays indicate quantities per resolution of the UNet model. $\{\}$ indicate an hyper-parameter search performed for each type of model (iCT, iCT-OT, iCT-GC ($\mu=0.5$)). }
\label{tab:hyperparameters_celeba}
\vspace{\baselineskip}
\begin{tabular}{ll}
\toprule
\textbf{Hyperparameter} & \textbf{Value} \\
 \midrule
batch size & $128$ \\
image resolution & $64$ \\
training steps & \num{150000} \\
learning rate & {0.00008} \\
optimizer & lion \\
$s_0$ & $10$ \\
$s_1$ & $1280$ \\
$\rho$ & $7$ \\
$\sigma_{0}$ & $0.002$ \\
$\sigma_{1}$  & $80$ \\
network architecture & SongUNet \\ & (from \cite{karras2022edm} implementation) \\
model channels & $128$ \\
dropout & $\{0., [0.,0.,0.2,0.2]\}$ \\
num blocks & $[3,3,4,5]$ \\
embedding type & positional \\
channel multiplicative factor &  $[1,2,2,2]$ \\
attn resolutions & $\emptyset$ \\
\bottomrule
\end{tabular}
\end{table}

\begin{table}
\centering
\caption{Hyperparameters for ImageNet-1k. Arrays indicate quantities per resolution of the UNet model. $\{\}$ indicate an hyper-parameter search performed for each type of model (iCT, iCT-OT, iCT-GC ($\mu=0.5$)). }
\label{tab:hyperparameters_imagenet}
\vspace{\baselineskip}
\begin{tabular}{ll}
\toprule
\textbf{Hyperparameter} & \textbf{Value} \\
 \midrule
batch size & $512$ \\
image resolution & $32$ \\
training steps & \num{150000} \\
learning rate & {0.00008} \\
optimizer & lion \\
$s_0$ & $10$ \\
$s_1$ & $1280$ \\
$\rho$ & $7$ \\
$\sigma_{0}$ & $0.002$ \\
$\sigma_{1}$  & $80$ \\
network architecture & SongUNet \\ & (from \cite{karras2022edm} implementation) \\
model channels & $128$ \\
dropout & $\{0., [0.,0.,0.2,0.2]\}$ \\
num blocks & $[3,5,7]$\\
embedding type & positional \\
channel mult & $[1,1,2]$ \\
attn resolutions & $[16]$ \\
\bottomrule
\end{tabular}
\end{table}

\textbf{Details on computational ressources} As mentioned in the paper, the image dataset experiments have been conducted on NVIDIA A100 40GB GPUs.

\end{document}